\documentclass{article}
\usepackage{CJKutf8}
\usepackage{xcolor}
\usepackage{graphicx}
\usepackage{arxiv}
\usepackage[T1]{fontenc}    
\usepackage{hyperref}       
\usepackage{url}            
\usepackage{booktabs}       
\usepackage{amsfonts}       
\usepackage{amsmath}
\usepackage{nicefrac}       
\usepackage{microtype}      
\usepackage{lipsum}
\usepackage[utf8]{inputenc}
\usepackage{multirow}
\usepackage{graphicx}
\usepackage{tabularx}
\usepackage{makecell}
\usepackage{arydshln}

\newcommand{\seq}[1]{\boldsymbol{#1}}

\DeclareMathOperator*{\argmax}{argmax}

\title{TranSmart: A Practical Interactive Machine Translation System}
\author{
  Guoping Huang, Lemao Liu, Xing Wang, Longyue Wang, \\
  \textbf{ Huayang Li, Zhaopeng Tu, Chengyan Huang and Shuming Shi}\\
  \\
  Tencent AI Lab\\
 \texttt{\{transmart,donkeyhuang,redmondliu,zptu,shumingshi\}@tencent.com}
}

\begin{document}
\maketitle
\begin{CJK}{UTF8}{gbsn}

\begin{abstract}
Automatic machine translation is super efficient to produce translations yet their quality is not guaranteed. 
This technique report introduces TranSmart, a practical human-machine interactive translation system that is able to trade off translation quality and efficiency. Compared to existing publicly available interactive translation systems, TranSmart supports three key features, word-level autocompletion, sentence-level autocompletion and translation memory.
By word-level and sentence-level autocompletion, TranSmart allows users to interactively translate words in their own manners rather than the strict manner from left to right. In addition, TranSmart has the potential to avoid similar translation mistakes by using translated sentences in history as its memory. 
This report presents major functions of TranSmart, algorithms for achieving these functions, how to use the TranSmart APIs, and evaluation results of some key functions. TranSmart is publicly available at its homepage~\footnote{ \url{https://transmart.qq.com}}.

\end{abstract}


\section{Introduction}

Recent years have witnessed a breakthrough in automatic machine translation, thanks to the advances in neural machine translation (NMT)~\cite{Vaswani:2017:NeurIPS,gehring2017convolutional}. The key idea in NMT is an encoder-decoder framework where a source sentence is represented by an encoder network and a target sentence is generate by an decoder network~\cite{sutskever2014sequence,bahdanau2014neural}. By using a large-scale bilingual corpus, both encoder and decoder networks consisting of massive parameters can be trained sufficiently enough to yield excellent generalization ability on unseen sentences. As a result, NMT delivers state of the art performance in many machine translation benchmarks~\cite{ondrej2017findings,barrault2019findings,wu2020tencent}.

\begin{figure}
	\centering  
	\includegraphics[width=0.5\textwidth]{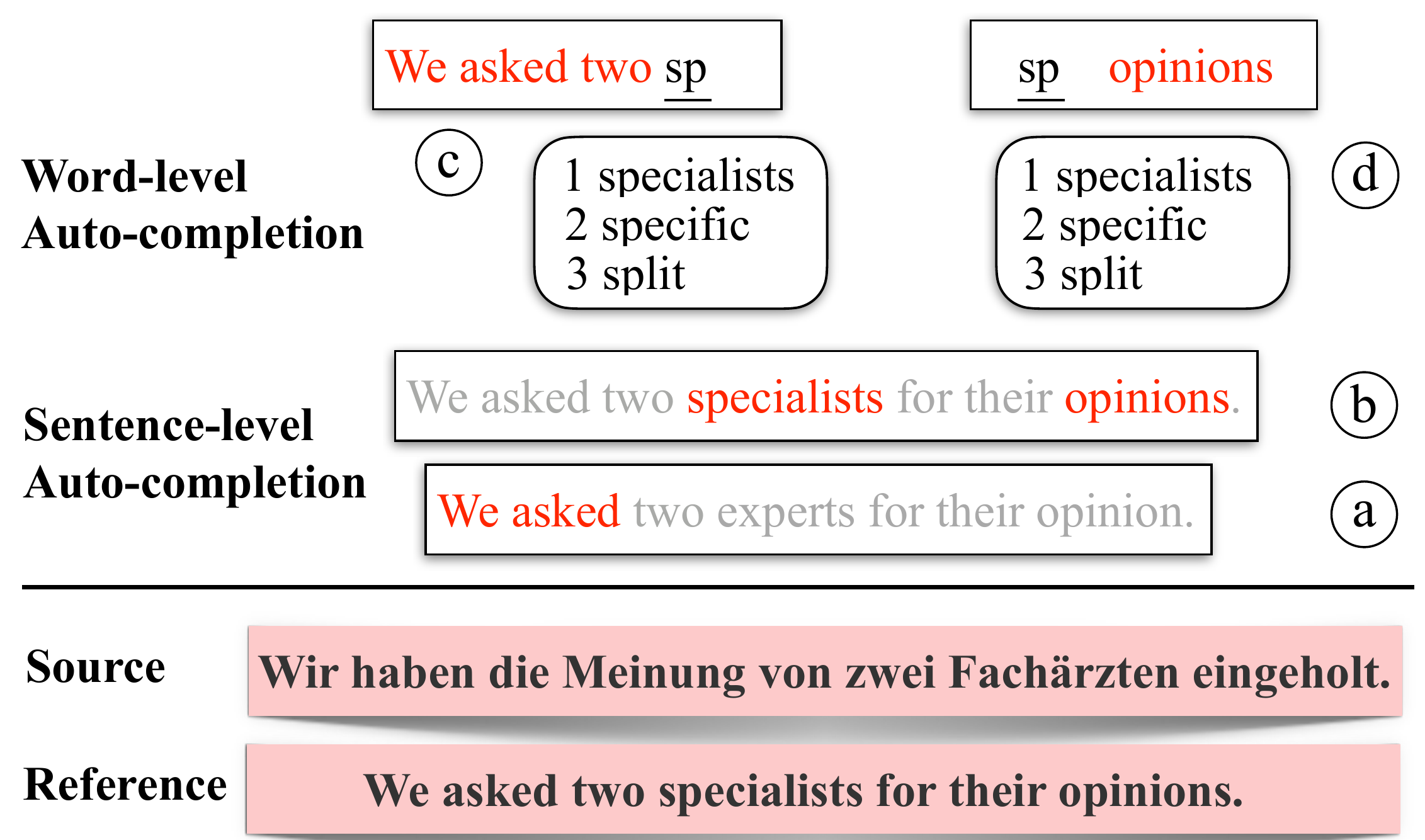}  
	\caption{Autocompletion comparison between conventional IMT (a \& c) and TranSmart (b \& d). For conventional IMT, autocompletion at sentence level (a) or word level (c) performs with the left-to-right direction; while TranSmart enables any directions as users enjoy for both sentence level (b) and word level autocompletion (d). Red words denotes those validated by users while gray words are automatically completed.}
	\label{fig:overview}  
\end{figure}
To the date, however, even a state-of-the-art NMT system is incapable of meeting the strict requirements in some translation scenarios, where human translation is still intensively involved even if it is inefficient and costly~\cite{peris2017interactive,weng2019correct,zhao+:2020:balancing}. Therefore, interactive machine translation (IMT)~\cite{plitt2010productivity,green2014predictive}, which is believed to better trade off translation quality and efficiency, has drawn increasing attention~\cite{knowles2016neural,grangier+auli:2018,wang+:2020:touch}: At each time a user corrects a translation, the machine automatically generates a new translation based on the human corrections until the translation process is finished~\cite{foster1997target}. Compared to automatic machine translation, an interactive machine translation system includes more components and it is more complicated in essence. During the past decade, there were a few IMT systems available, for instance, TransType~\cite{langlais2000transtype}, CASMACAT~\cite{alabau2014casmacat}, and LILT~\footnote{\url{www.lilt.com}.} which are either on top of statistical machine translation~\cite{koehn-etal-2003-statistical,chiang2005hierarchical,koehn2009statistical} or the advanced NMT. All these systems share a common characteristic: they perform translation in an incremental manner from left to right and naturally require human translators to follow this strict manner, as shown in Figure 1 (a \& c), no matter what translation manner of translators is.

In this report, we describe a new human-machine interactive translation system, TranSmart, which was first made publicly accessible in 2018. Compared with other IMT systems, one advantage of TranSmart is its flexibility in the sense that users can employ any translation manners to interact with the machine rather than the incremental manner from left to right. Therefore, if some users prefer to translate difficult words first and then easy words, TranSmart is able to conduct interactive translation in their own manners, as shown in Figure 1 (b \& d). Moreover, TranSmart includes an NMT engine augmented with translation memory, and it is very helpful to translate a document where similar translation errors may occur for an automatic machine translation engine. To the best of our knowledge, TranSmart is the first interactive neural machine translation (INMT) system which makes use of translation memory. 

Specifically, TranSmart provides three key features as follows:
\begin{itemize}
    \item Word-level autocompletion: At word level, in order to input a correct word, users do not need to type all characters of this word from scratch but a few characters instead, and then the system can automatically complete this word. Unlike other INMT systems, the corrected word is not necessary to be adjacent to the translation prefix. 
    \item Sentence-level autocompletion: At sentence level, users do not need to translate all words from scratch but provide some of them (for instance, some difficult words), and then the system can automatically complete the translation based on user provided words. Different from other INMT systems, user provided words can be discontinuous. 
    \item Translation-memory-augmented NMT: The system is able to re-use translation results from users through translation memory and offers better translation. As there may be several similar sentences in a document to be translated, the system can efficiently avoid the occurrence of similar translation errors thanks to the knowledge from translation memory. 
\end{itemize}
By repeatedly leveraging these key features, TranSmart is able to interactively collaborate human and a machine to generate a high-quality translation in an efficient way.
In addition, TranSmart provides some extended features, including terminology translation, bilingual sentence examples, document translation, tag-preserving translation, and image translation.
In the remaining part of this report, we first present our implementation of the key modules of TranSmart. Then the TranSmart API is briefly introduced. Finally, the effectiveness of some functions is demonstrated through empirical experiments.

\section{System Features}

At high level, TranSmart contains three key features as well as several extended features as shown in the left side of Figure~\ref{fig:overview}. This section describes the basic ideas of these features, whose detailed implementations will be given in the next section. 

\begin{figure}
	\centering  
	\includegraphics[width=0.7\textwidth]{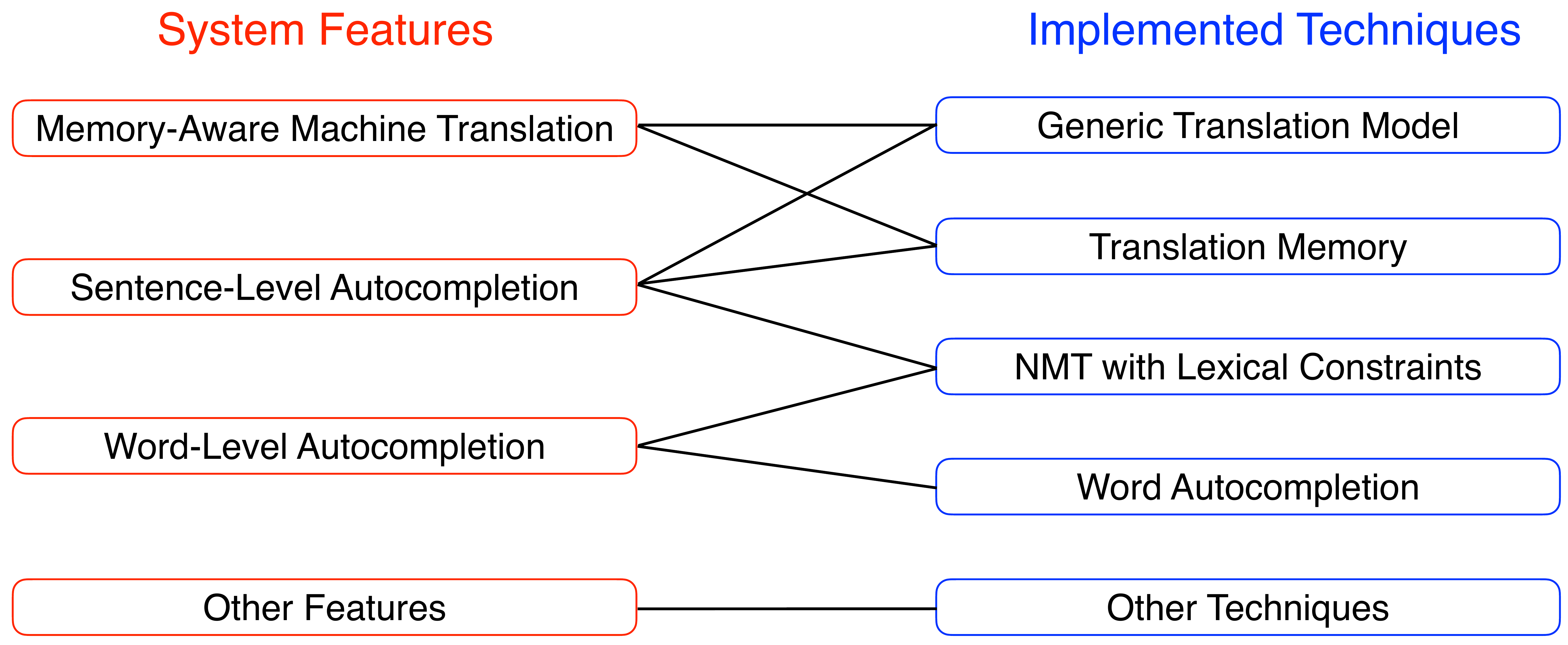}  
	\caption{Overview of the TranSmart system.}
	\label{fig:overview}  
\end{figure}

\subsection{Key Features}
\paragraph{Word Level Autocompletion} Given a source sentence, a translation context consisting of translation pieces, and a human-typed character sequence, word level autocompletion aims to predict a target word which is compatible with the typed sequence. With the help of this feature, if a user is expected to type a target word to correct a translation, it is not necessary to manually type all the characters but only some of them. This feature is inspired from recent advances~\cite{chen2000new} in input methods from monolingual scenario which are designed to improve the efficiency of human input. 

\paragraph{Sentence Level Autocompletion}
Given a source sentence and a translation context, sentence-level autocompletion aims to generate a complete translation for the source sentence on top of the context. The translation context can be human typed words (with or without word level autocompletion), or human-edited translation pieces from a translation generated by the system. With this feature, a user does not need to manually translate all words from scratch but some of them which are usually difficult to the system, and then the system tries to complete the translation by generating the remaining words automatically. 

\paragraph{Memory-Aware Machine Translation} Memory-Aware Machine Translation aims to generate high-quality translation by making use of translation memory. Users can provide their own domain-specific translation data to our system as the memory or we can use our bilingual training corpus as the memory. In addition, after users finish translating a sentence, we have a mechanism to accumulate their translation history into the memory. In this way, our system has the ability to avoid the same translation errors occurring multiple times, and this feature is useful in translating a document where there are similar sentences.  

The above three key features can be used as atomic operations, which can be applied multiple times to interact between the user and machine for a translation task. For example, by repeatedly leveraging word level autocompletion and sentence level autocompletion, TranSmart can generate a complete translation for a source sentence with high-quality, by using the memory-aware translation engine where translation memory is accumulated from translation history of users. It is worth noting that the translation pieces in the translation context may be discontinuous, and thus our word level and sentence level autocompletion is more general and flexible than those in existing INMT systems. 

\subsection{Extended Features}
\paragraph{Document Translation} This feature is used to translate a formatted document in a source language to a corresponding formatted document in a target language. It supports many popular formats including \texttt{TXT, HTML, XML, MARKDOWN, PDF, DOCX, PPTX} and \texttt{XLSX}. To this end, it generally performs two steps as follows. First, it parses the input formatted document into a text document consisting of sentences with several tags. Each tag may indicate some structural information such as a paragraph or a font. 
For example, a sentence in the formatted document may be "\texttt{<style text-fill="red">Forrest Gump</style> is a 1994 American drama film}", where "\texttt{<style text-fill="red">Forrest Gump</style>}" indicates the phrase \texttt{Forrest Gump} is with red color as its format. Second, it translates each tagged sentence by the automatic translation engine with a specially designed technique, which is named tag translation. Tag translation is crucial to document translation and we will present its challenges and our solution in the next section. 

\paragraph{Image Translation} This feature aims to translate text contained in an image file in a source language to a text document in a target language. It supports many popular image formats such as \textit{JPG} and \textit{PDF}. Generally, it is implemented by a pipeline procedure consisting of two steps as follows. First, it employs an external OCR toolkit to extract a text document where any content beyond text is ignored. Second, it uses our translation engine to translate all sentences in the text document one by one. The first step, which is called text extraction from image, is critical to image translation and we will present its challenge and the technique to address in next section. 

\paragraph{Terminology Translation} We collect more than 3 million of Chinese-English terminologies from websites. However, this corpus contains a large amount of noises, including non-terminology, unaligned, inconsistent format. We filter non-terminology words by considering the word frequency due to the long-tail property of terminology. More specifically, we built a phrase table from a large-scale parallel data and then extract high-frequency phrase pairs as a non-terminology list. Finally, we filter noises by comparing the stop list and collected data~\cite{wang2014systematic}. Furthermore, we employ our in-house filtering scripts~\cite{wang2014combining} to filter unaligned terms according to various features such as length ratio and language identification. As a result, we obtained a clean version of terminology corpus that contains around 2 million terms.

\paragraph{Bilingual Examples} The input sentence is used to retrieve bilingual examples from the corresponding retrieval repository. We selected and used more than 200M bilingual sentences to build the retrieval repository. The three most similar bilingual examples are displayed to help users to translate the input sentence.

\section{Implemented Techniques}
\subsection{Generic Translation Model}

\paragraph{Translation Model}
We implemented the generic translation model on top of the Transformer architecture~\cite{Vaswani:2017:NeurIPS}. 
To balance the translation performance and inference efficiency, we used a 24-layer encoder and a 6-layer decoder, whose hidden size is 1024.
We trained the translation model on our in-house data after the following data manipulation methods, which consists of 200 millions of Chinese-English sentence pairs. 
We followed~\cite{Ott:2018:WMT} to train models with batches of approximately 460k tokens, using Adam~\cite{kingma2015adam} with $\beta_{1}=0.9$, $\beta_{2}=0.98$ and $\epsilon=10^{-8}$.  


\begin{figure*}[h]
    \centering
    \includegraphics[width=0.7\textwidth]{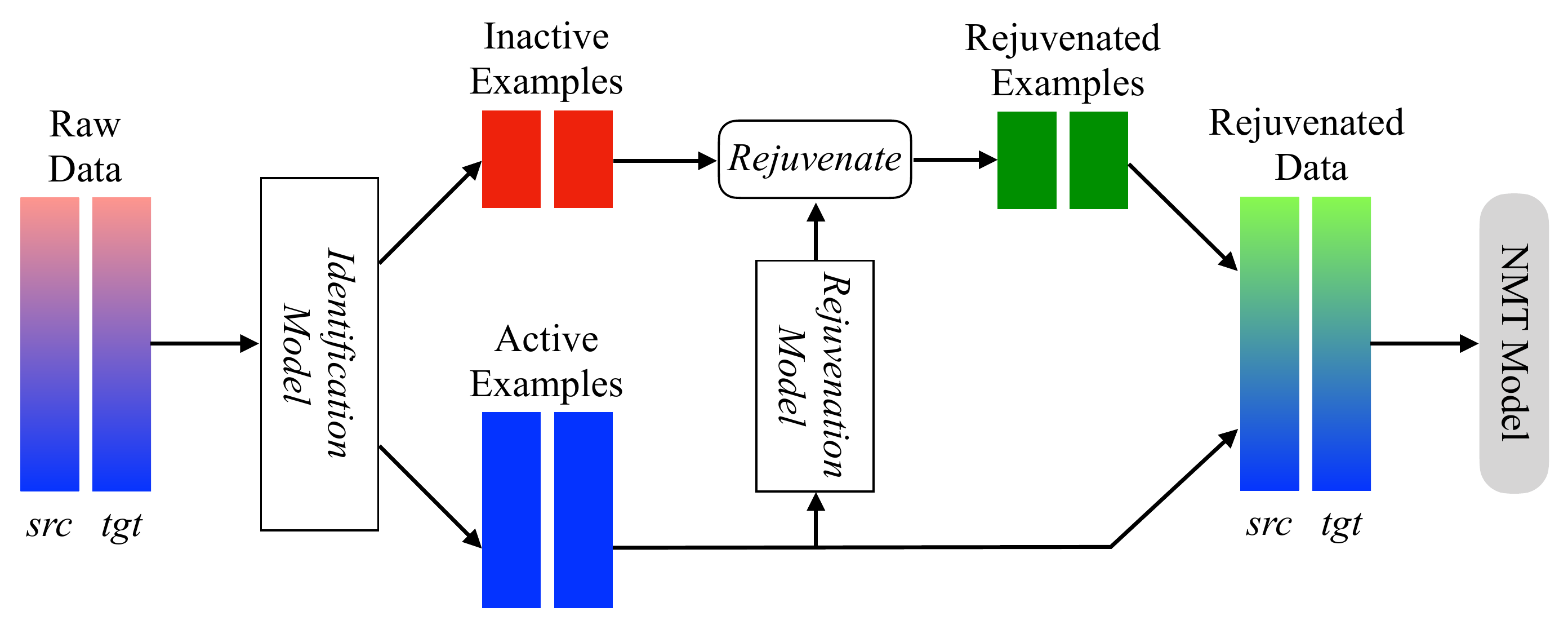}
    \caption{The framework of data rejuvenation, which consists of two models. The {\em identification model} identifies inactive examples from the original training data, which is then rejuvenated by the {\em rejuvenation model}. The rejuvenated examples along with the active examples are used together to train the NMT model.} 
    \label{fig:data-rejuvenation}
\end{figure*}

\paragraph{Data Rejuvenation}
Large-scale parallel datasets lie at the core of the recent success of NMT models. However, the complex patterns and potential noises in the large-scale data make training NMT models difficult. 
We introduce \emph{data rejuvenation} to improve the training of NMT models on large-scale datasets by exploiting inactive examples~\cite{Jiao:2020:EMNLP}. The proposed framework consists of three phases, as shown in Figure~\ref{fig:data-rejuvenation}. 
First, we train an {\em identification model} on the original training data to distinguish inactive examples and active examples by their sentence-level output probabilities.
Then, we train a {\em rejuvenation model} on the active examples to re-label the inactive examples with forward-translation.  Finally, we combined the rejuvenated examples and the active examples as the final bilingual data.

\paragraph{Data Augmentation}
Although we have large-scale parallel data, there are limited amount parallel data for some specific domains. Data augmentation methods (e.g. self-training and back-translation) are a promising way to alleviate this problem by augmenting model training with synthetic parallel data. The common practice is to construct synthetic data based on a randomly sampled subset of large-scale monolingual data, which we empirically show is sub-optimal. In response to this problem, we improve the sampling procedure by selecting the most informative monolingual sentences to complement the parallel data. To this end, we compute the uncertainty of monolingual sentences using the bilingual dictionary extracted from the parallel data. Intuitively, monolingual sentences with lower uncertainty generally correspond to easy-to-translate patterns which may not provide additional gains. Accordingly, we design an uncertainty-based sampling strategy~\cite{Jiao:2021:ACL} to efficiently exploit the monolingual data for self-training, in which monolingual sentences with higher uncertainty would be sampled with higher probability. 

\subsection{General Word-level Autocompletion}
\begin{figure*}[t]
  \begin{center}
      \includegraphics[width=0.65\columnwidth]{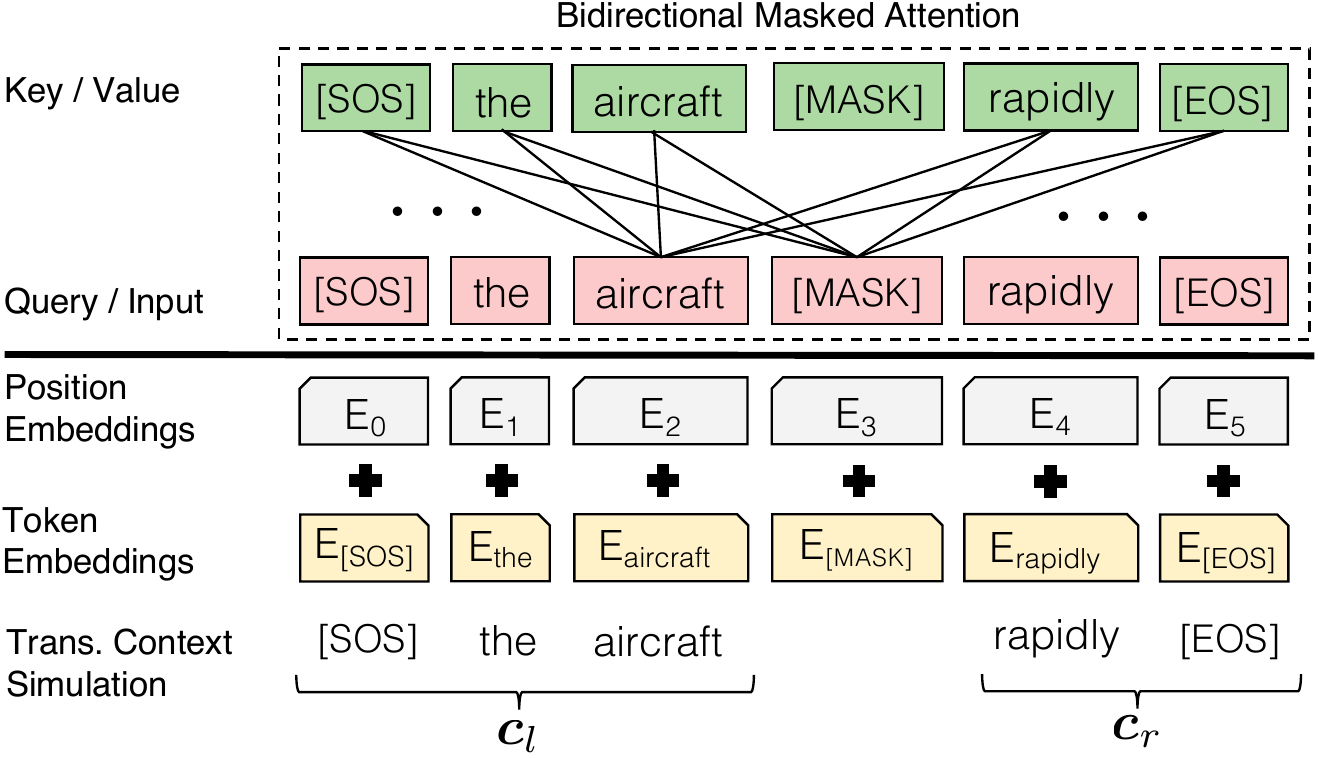}
      \caption{The input representation of our model and architecture of Bidirectional Masked Attention. The input embeddings are the sum of the token embeddings and position embeddings. \texttt{[MASK]} represents the potenial target word in this translation context. \label{fig:attention}}
  \end{center}
\end{figure*}
\paragraph{Task Definition}
Word-level autocompletion aims to complete the target word based on human typed characters for a given a source sentence and translation context. Previous studies have explored word-level autocompletion task, but they either do not take into account translation context~\cite{huang2015new} or they require the target word to be the next word of the translation prefix~\cite{langlais2000transtype, santy-etal-2019-inmt}, which limit its applications in real-world scenarios such as post-editing~\cite{vasconcellos1985spanam}. To this end, we propose a general word-level autocompletion task which can be applied to more general scenarios.

Suppose $\seq{x}$=$(x_1, x_2, \dots, x_m)$ is a source sequence, $\seq{s}$=$(s_1, s_2, \dots, s_k)$ is a sequence of human typed characters, and translation context is denoted by $\seq{c}$=$(\seq{c}_l, \seq{c}_r)$, where $\seq{c}_l$=$(c_{l,1}, c_{l,2}, \dots, c_{l,i})$ and $\seq{c}_r$=$(c_{r,1}, c_{r,2}, \dots, c_{r,j})$. The translation pieces $\seq{c}_l$ and $\seq{c}_r$ are on the left and right hand side of $\seq{s}$, respectively. Formally, given a source sequence $\seq{x}$, typed character sequence $\seq{s}$ and a context $\seq{c}$, the \textit{general word-level autocompletion} 
(GWLAN) task aims to predict a target word $w$ which is to be placed in the middle between $\seq{c}_l$ and $\seq{c}_r$ to constitute a partial translation. Note that $\seq{c}_l$ or $\seq{c}_r$ may be empty in some scenarios.

\paragraph{Methodology}
Given a tuple $(\seq{x}, \seq{c}, \seq{s})$, our approach decomposes the whole word autocompletion process into two parts: model the distribution of the target word $w$ based on the source sequence $\seq{x}$ and the translation context $\seq{c}$, and find the most possible word $w$ based on the distribution and human typed sequence $\seq{s}$. 

In the first part, we propose a word prediction model (WPM) to define the distribution $p(w|\seq{x}, \seq{c})$ of the target word $w$. We use a single placeholder \texttt{[MASK]} to represent the unknown target word $w$, and use the representation of \texttt{[MASK]} learned from WPM to predict it. Formally, given the source sequence $\seq{x}$, and the translation context $\seq{c}=(\seq{c}_l$, $\seq{c}_r)$, the possibility of the target word $w$ is:
\begin{equation}\label{eq:objective}
 P(w|\seq{x}, \seq{c}_l, \seq{c}_r;\theta) = \mathrm{softmax} \left(\phi (h) \right)[w]
\end{equation}
where $h$ is the hidden representation of decoding state with respect to \texttt{[MASK]}, $\phi$ is a linear network that projects the hidden representation $h$ to a vector with dimension of target vocabulary size, and $\mathrm{softmax}(d)[w]$ takes the component regarding to $w$ after the softmax operation. Our model has a source encoder and a cross-lingual encoder.  The source encoder of WPM is the same as the Transformer encoder, which is used to encode the source sequence $\seq{x}$. The output of source encoder is passed to the cross-lingual encoder later. The cross-lingual encoder is similar to the Transformer decoder, while the only difference is that we replace the auto-regressive attention layer by a bidirectional masked attention (BMA) module, which is shown in 
 Figure \ref{fig:attention}.


Suppose $\seq{s}$ denotes a human typed sequence of characters, in the second part, we predict the best word according to the constrained optimization:
\begin{equation}
{\argmax}_{w \in \mathcal{V}(\seq{s})} P(w\mid \seq{x}, \seq{c}; \theta)
\end{equation}
\noindent where $\mathcal{V}(\seq{s})$ denotes a set of target words, whose element satisfies the sequences of $\seq{s}$, for example, $\seq{c}$ is a prefix of $w$.  
More details can be found in \cite{li2021gwlan}.

\paragraph{Data Generation}
For training and evaluating GWLAN models above, firstly we should create a large scale dataset including tuples of $(\seq{x}, \seq{s}, \seq{c}, w )$. Ideally, we may hire professional translators to manually annotate such a dataset, but it is too costly in practice. We instead propose to automatically construct the dataset from parallel datasets. 

Assume we are given a parallel dataset $ \{(\seq{x}^i, \seq{y}^i)\}$, where $\seq{y}^i$ is the reference translation of $\seq{x}^i$. 
We automatically construct the data $\seq{c}^i$ and $\seq{s}^i$ by randomly sampling from $\seq{y}^i$. Specifically, we first sample a word $w=\seq{y}^i_k$, and sample two spans $[a_l, b_l]$ and $[a_r, b_r]$ such that $0\le a_l \le b_l \le k$ and $k+1\le a_r \le b_r \le |\seq{y}^i|$, leading to $\seq{c}_l = \langle y^i_{a_l}, \cdots, y^i_{b_l-1}\rangle$ and $\seq{c}_r = \langle y^i_{a_r}, \cdots, y^i_{b_r-1}\rangle$. It is worth mentioning that $\seq{c}_l$ or $\seq{c}_r$ may not be adjacent to $y^i_k$. In addition, we randomly sample a character sequence as $\seq{s}^i_k$ for $y^i_k$ as follows: for languages like English and German, $\seq{s}^i_k$ is a character prefix of $y^i_k$; for languages like Chinese, $\seq{s}^i_k$ consists of the first phonetic symbol of each character in $y^i_k$. 
In this way, we can obtain a collection of $\{(\seq{x}^i, \seq{c}_l, \seq{c}_r, \seq{s}^i_k, y^i_k)\mid \forall i, \forall k\}$, which can be divided into training, validation and test datasets for the GWLAN task.

\subsection{Sentence-level Autocompletion by Lexical Constraints}
In our human-machine interactive translation scenario, human translators may pre-specify some constraint words (or lexical constraints) and our system requires to output a high-quality translation with knowledge from these pre-specified constraints. For example, the constraints can be typed by human translators through word level autocompletion in Section 3.2, and they can be a translation prefix corrected by translators, or a post-edited partial translation. It is worth noting that the constraints are not necessary to be continuous~\cite{cheng2016primt}, unlike \cite{wuebker2016models,knowles2016neural}. In TranSmart, we implement two different approaches to incorporate these constraints into NMT.
The first one relies on constrained decoding which requires the output translation to include constraints in a hard manner; whereas the second one makes use of them in a soft manner, i.e., the output translation may not include some of constraints. 

\paragraph{Constrained Decoding}

Constrained decoding is essentially a constrained optimization problem, and it aims to search the translation which satisfies some constraints and is with the best model score. Formally, suppose $P(\seq{y}\mid \seq{x})$ is a translation model, and $\seq{c}$ denotes a set of constraint words. 
$$
\argmax_{\seq{y}\in \mathcal{Y}(\seq{c})} P(\seq{y}\mid \seq{x})
$$
\noindent where $\mathcal{Y}(\seq{c})$ denotes a set of translation hypotheses which include all constraint words in $\seq{c}$.
To  address  this  constrained  optimization  problem, \cite{hokamp2017lexically} proposed a grid beam search algorithm (GBS). This algorithm  maintains  a  beam  along  two dimensions, where one is the  length  of  the  hypotheses and  the  other  is  the  number words in all constraints. Thus, its complexity is linear to the number of the constraint words, which is inefficient in our interactive scenario. \cite{post2018fast} presented an improved algorithm by dynamic beam allocation (DBA) which allows hypotheses in a beam to contain different number of constraint words. In our experiments, this improved algorithm is indeed more efficient in decoding speed but its translation quality is worse than grid beam search. To this end, we propose a variant of grid beam search algorithm to achieve a trade-off between quality and efficiency. 

In our interactive scenario, we observe that constraint words in $\seq{c}$ exhibit two characteristics: 
some of constraints in $\seq{c}$ are consecutive to form a translation piece, especially when the size of $\seq{c}$ is very large; and 
$\seq{c}$ provided by human translators is placed in a fixed order such that the output translation should contain $\seq{c}$ in the same order.
Based on the observation, we organize a beam along the length of hypotheses and the number of constraint pieces rather than constraint words.  Thanks to the fixed order of pieces, any translations with the same number of pieces will naturally contain the same number of constraint words and thereby they can be fairly pruned in terms of model scores, similar to the grid beam search algorithm. Consequently, the complexity of this algorithm is linear to the number of pieces rather than the number of constraints, which leads to substantial speedup in practice because constraint pieces usually are long enough. 

\paragraph{NMT with Soft Constraints} 
Unlike constrained decoding, we propose a new approach to make use of pre-specified constraints in a soft manner. The key is to treat the constraints as an external memory, integrate the memory into the standard Seq2Seq decoder~\cite{sutskever2014sequence,bahdanau2014neural} and then train the memory-augmented NMT from a dataset such that it learns to constrain the output.
Specifically, given $\seq{x}$, $\seq{c}$ and a translation prefix $\seq{y}_{<t}$, we generate a target word $\seq{y}_t$ according to
$P\left( {y}_t \mid \seq{x}, \seq{y}_{<t}, \seq{c}; \theta \right)$. 
Our model includes three components: 
the encoder of vanilla NMT model which encodes $\seq{x}$ into a sequence of hidden vectors; the Constraint Memory Encoder (CME) that encodes the constraints $\seq{c}$ into another sequence of hidden vectors $E(\seq{c})$, 
i.e., the constraint memory; and the Constraint Memory Integrator (CMI), which integrates the constraint memories into the decoder network to generate the next token $y_t$. 
To train the memory-augmented model, we create a dataset from available bilingual corpus.  We randomly sample some rare words from the reference $\mathbf{r}$ as constraints, since rare words are more difficult to translate than other words \cite{grangier+auli:2018}.

In inference, 
our decoding is an unconstrained optimization problem:
$$
\argmax_{\seq{y}} P(\seq{y}\mid \seq{x}, \seq{c}; \theta)
$$
To approximately solve the above optimization problem, we employ the standard beam search algorithm which is similar to the decoding in our baseline model. Although we need additional overheads to handle the constraint memories, this is negligible to the time consuming of the decoding in NMT. Thus, its search is as efficient as the search algorithm of the standard NMT.
In addition, unlike constrained decoding, the method does not force a feasible $\seq{y}$ to include all constraints in $\seq{c}$. Hence, if the constraints in $\seq{c}$ include some noises (i.e., spelling mistakes), it has the potential to avoid copying these noises. More details about this function can be found in~\cite{li+:2020:neural}.

\subsection{Graph based Translation Memory}
\begin{figure*}[t!]     
\centering                                     
\includegraphics[width=1.\textwidth]{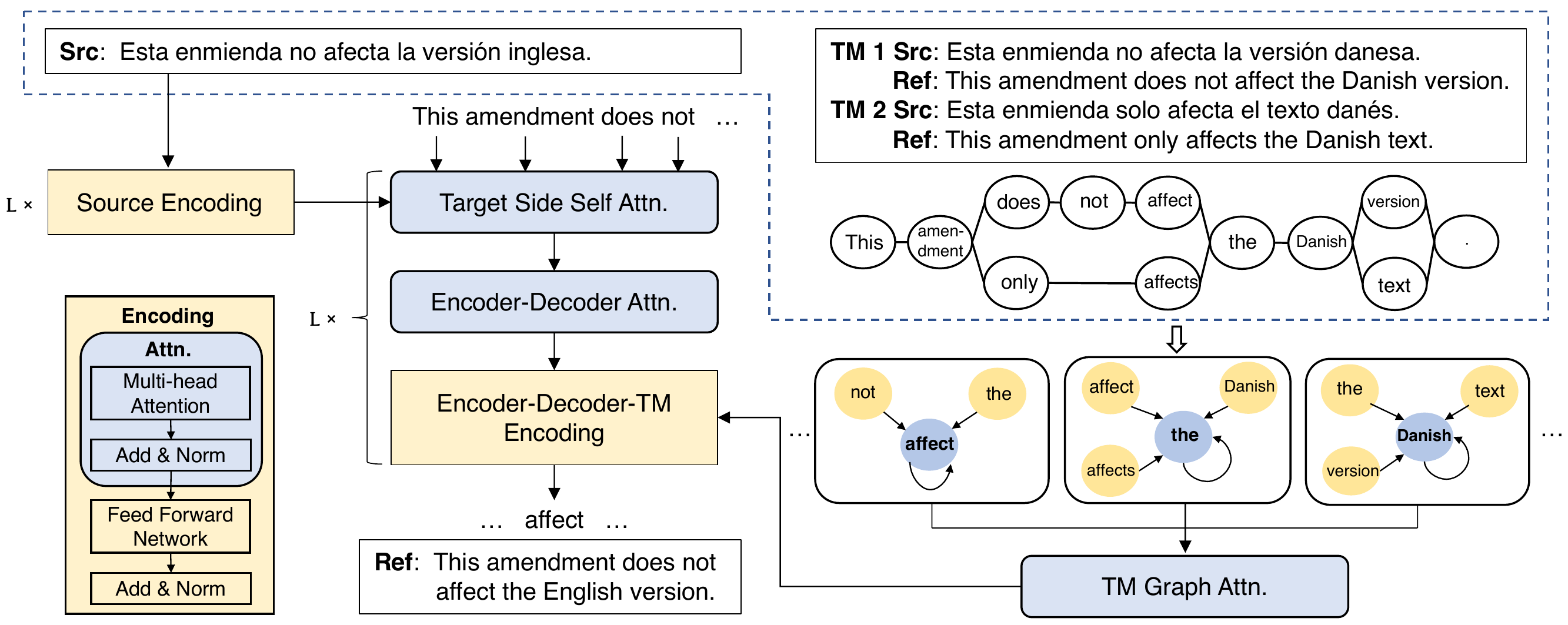}
\caption{The architecture of the proposed NMT with graph based TM. 1) Graph representation - The part in the dashed box is a concrete example of the graph representation of a TM.  2) Model architecture - The part outside of the dashed box shows the core components of the model architecture. }  
\vspace{-5mm}
\label{fig:topdown}
\end{figure*}

The basic idea of translation memory based MT is to translate an input source sentence by using the translations of the source sentences which are similar to the input one, which is related to domain adaptation~\cite{chu2017empirical,wang2017instance,chu-wang-2018-survey}. 
Suppose we are given a translation memory (TM) for a source sentence which is a list of bilingual sentence pairs.  Generally, there are two ways to improve translation models with translation memory: training model parameters with augmented data (i.e., memory)~\cite{liu+:2012:locally,li2016one,farajian2017multi} and summarizing knowledge from translation memory to augment MT decoder~\cite{gu+:2018,zhang+:2018,he+:2019:word}. For the latter idea, a typical solution to represent a TM is to encode each word in both the source and target sides by a neural memory\cite{gu+:2018}. Unfortunately, since a word even a phrase may appear repeatedly in a TM, redundant words are encoded multiple times, leading to a large memory network as well as considerable computation. To address this issue, we propose an effective approach to representing a TM with a compact graph structure, which is further used to enhance the translation model. The model structure is illustrated in Figure ~\ref{fig:topdown} and its more details can be found in \cite{xia+:2019:graph}.

\paragraph{Graph Representation of TM}
It is observed that most source words in the TM also appear in the input sentence and have already been represented by the encoder. In addition, we believe that those words in the TM yet beyond the source side may not be informative to translate the input sentence itself. Therefore, in our proposed model, we directly ignore the source sentences of the TM and only represent the target side. In addition, instead of sequentially encoding target sentences in a TM, we pack them into a compact graph such that some words in different sentences may correspond
to the same node in the graph, which is inspired by the notion of lattice or hypergraph in statistical machine translation \cite{koehn:2009:book,dyer2010cdec}. 
To this end, we convert the target side in a TM into a confusion network by using the algorithm proposed by \cite{mangu+:1999} and \cite{mangu+:2000}. 

\paragraph{NMT with Graph based TM}
The graph based TM is further used to enhance the Transformer architecture. 
Generally, the enhanced Transformer shares the similar architecture as Transformer but with two major differences in encoding and decoding phases. 

In the encoding phrase, besides encoding the input sequence, the proposed model also encodes the graph by using $L$ layers of networks in a similar fashion to the encoding of the input. Specifically, we firstly use a multi-head attention to encode each node in the TM graph where the query is the corresponding node, and key-value pairs are obtained from its first-order neighborhood nodes inspired by graph attention~\cite{velivckovic+:2018}. Then we apply other sub-layers to the resulting vector obtained by the multi-head attention layer, which contain a residual layer, a feed-forward layer and a layer normalization. In this way, we can represent the graph into a list of vectors whose size is the same as the number of nodes in the graph.

In the decoding phase, similar to Transformer, the proposed model employs $L$ layers of networks but each layer includes sub-layers (i.e., multi-head attention, residual layer and layer normalization) to incorporate the list of vectors obtained from graph encoding, besides other six sub-layers. We place the extra three sub-layers nearest to the output of the decoder network in order to let the graph encoding fully influence the decoding process.

\subsection{Others}
\paragraph{Word Alignment}
Word alignment plays an important role in document translation for TranSmart. 
Since NMT is a blackbox model with massive parameters, previous work has made numerous efforts to induce word alignment from attention in NMT~\cite{liu+:2016:neural,li+:2018:target,chen+:2020:accurate} or other explanation methods~\cite{li+:2019:on,ding2019saliency,li+:2020:evaluating}. Other work improves alignment quality by building a word alignment model whose architecture is similar to NMT~\cite{chen2020maskalign}. Despite its success, statistical aligners~\cite{och2003systematic,dyer2013simple} are still respectful counterparts because of their training efficiency and alignment quality. Therefore, we employ statistical aligners to obtain word alignment. As TranSmart involves billion scale of bilingual sentences as its training data, the popular aligner GIZA++ can not train successfully due to memory consumption.
Instead, we re-implement an aligner based on HMM~\cite{vogel1996hmm} with an adaptive strategy to prune the word translation table: for each high-frequency word we allow more words to be its translations whereas we allow less for each low-frequency word. 

\paragraph{Tag Translation} 
Tag translation aims to translate a source language with tags into a target language with tags, and it serves as the key step for document translation. Since the standard translation engine are trained on top of bilingual sentences without tags, standard translation engines can not perform well on translating tagged sentences. In addition, there are no sufficient tagged bilingual sentences to train a customized tag translation engine. Therefore, we propose a simple post-processing approach based on word alignment as follows. First, we delete all tags from a tagged source sentence to obtain a tag-free sentence and then translate it into a target sentence by our default translation engine. Then we run our word aligner on both the tag-free sentence and its translation to obtain word level alignments. Second, we insert the corresponding tags from the source sentence into its translation. However, the second step is not straightforward because one tagged piece (or phrase) within the tagged source sentence may align to multiple pieces in the target side due to the essence of word alignment. We propose an algorithm based on dynamic programming to extract a piece-to-piece alignment: the piece-to-piece alignment is a one-to-one map between pieces in the source and target sides such that it makes the minimal violations according to the word alignment results. By using the piece-to-piece alignment, it is trivial to insert the tags from the source sentence into its target sentence.

\paragraph{Text Extraction from Image} 
Text Extraction from Image aims to extract the text content from an image file to form a text document in a source language while ignoring other content. Generally, this task is challenging due to two reasons. First,  a sentence in an image file may not explicitly contain a special symbol to indicate its end and several such sentences may actually constitute one sentence. More importantly, OCR may recognize many blocks of text content from an image file to some extent, but it is difficult to organize these text blocks into a text document which preserves the same sequential structure of text blocks as in the original image file. To tackle the first challenge, we develop a language model to detect the end symbol for a sentence without an end symbol. In addition, we design another model to detect whether several sentences without an end symbol should be combined to one sentence. To address the second challenge, we employ the position information of each block from the OCR toolkit and uses it as a signal to decide the sequential order of all blocks, which is critical to form the final text document. 

\paragraph{Discourse-Aware Translation}
Existing translation models usually translate a text by considering isolated sentences based on a strict assumption that the sentences in a text are independent of one another. However, disregarding dependencies across sentences will harm translation quality especially in terms of coherence, cohesion, and consistency~\cite{longyue2019discourse}. To response this problem, we adapt document-level NMT~\cite{wang2017exploiting} to document-level training. Specifically, we add bilingual documents and paragraphs to our training data, which can helps models to learn discourse knowledge from larger contexts~\cite{wang2016automatic,wang2020tencent}. This works well with the {\em Document Translation Feature} as stated in Section 2.2. Chinese is a pro-drop language, where pronouns are usually omitted when they can be inferable from the context. This leads to serious problems for Chinese-to-English translation models in terms of completeness and correctness. Thus, we recover missing pronouns in the informal domain of training data (e.g. conversations and movie subtitles) by leveraging our approaches~\cite{wang2016novel,wang2018translating,wang2018learning}.

\section{System Usage}

Two ways are available to use TranSmart. One way is to visit the website of TranSmart\footnote{\url{https://transmart.qq.com/index}}, which has a friendly interactive user-interface (UI). Another way is calling the TranSmart HTTP APIs. The HTTP APIs can be accessed by sending a JSON request to the service\footnote{\url{https://transmart.qq.com/api/imt}} via POST method. The major fields of the JSON request of the translation APIs are shown in Table \ref{tab:api_fields}. In this section, we will focus more on the latter way and introduce the usage of some selected features. 

\begin{table}[]
\begin{tabularx}{\textwidth}{|l|l|l|X|}
\hline
Field           & Field Type &  Req/Opt   & Description                                                                                                                                                                                           \\ \hline
header          & JSON       & Required & Information of API calls, including function name, token, and user name\footnote{The token and user name have to be registered in our account management system. Please consult with us for the details of registration process.}.                                                                                                                              \\ \hline
reference       & JSON       & Optional & Information of references, which includes two fields ``term\_lib'' and ``sentence\_lib''.                                                                                                             \\ \hline
type            & String     & Optional & The format of the given source sentence. The value should be one of plain, xml, html, and markdown. The default value is ``plain''.                                                                   \\ \hline
model\_category & String     & Optional & The model served for translation. The value should be one of slow, normal, and fast. The default value is ``normal''. Note that different models may have different translation speeds and qualities. \\ \hline
source          & JSON       & Required & Information of the source text, including a ``text\_list'' filed that contains a list of source sentences.                                                                                            \\ \hline
target          & JSON       & Required & Information of the target sentence.                                                                                                                                                                   \\ \hline
limit           & JSON       & Optional & Control the returns in the JSON response, such as the number of suggestions.                                                                                                                                                                       \\ \hline
reference\_list & JSON       & Optional & Information of translation memory. The translation memory has two types: ``term'' and ``sent''.                                                                                                       
\\ \hline
\end{tabularx}
\caption{\label{tab:api_fields} Major fields of the request JSON object when calling TranSmart API. Req/Opt denotes the field is required or optional.}
\end{table}

\subsection{Word And Sentence Level Autocompletion}
\begin{figure*}[]     
\centering                                     
\includegraphics[width=1.\textwidth]{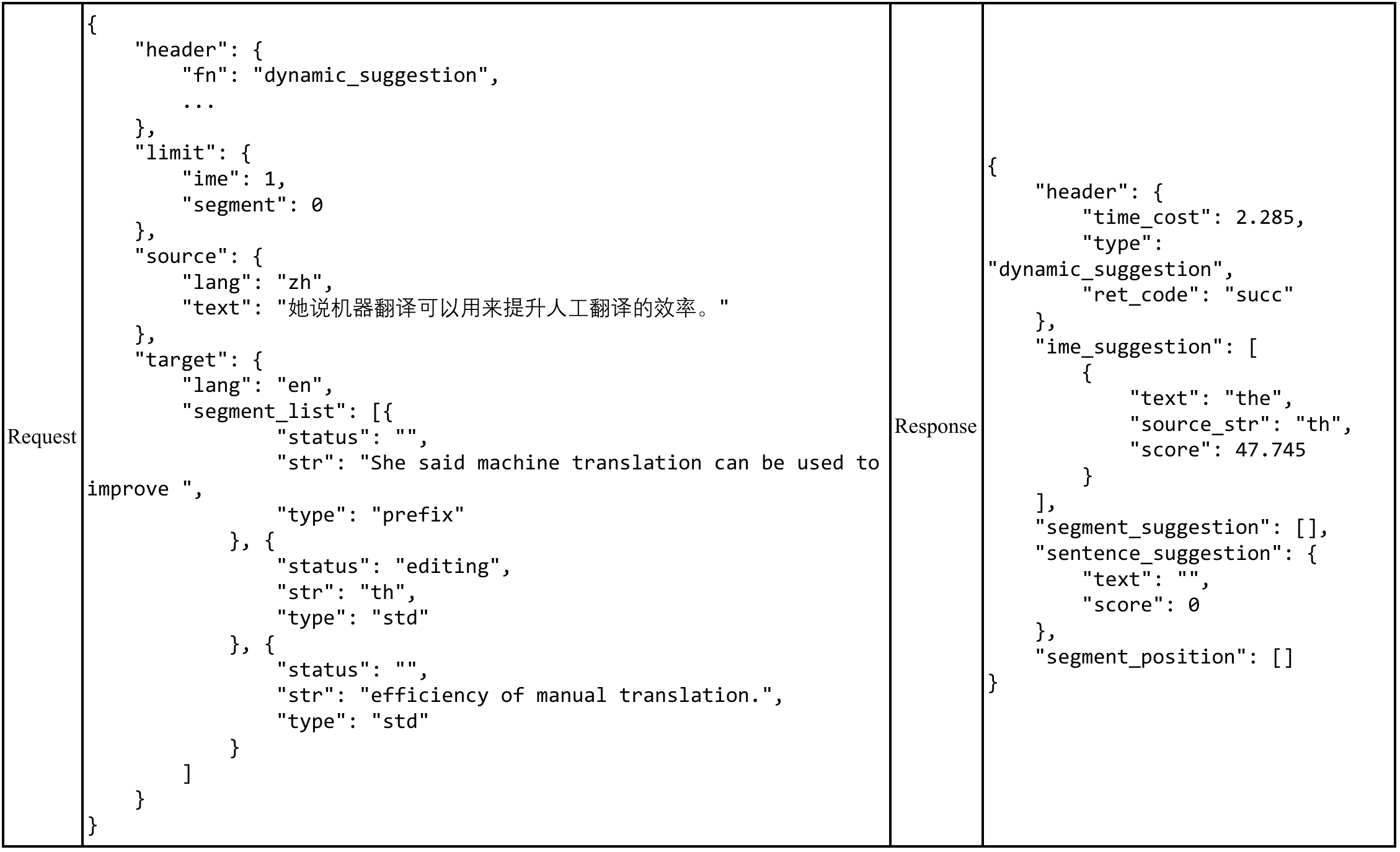}
\caption{Example of the word-level dynamic suggestion.}  
\label{fig:dynamic_suggestion_ime}
\end{figure*}

\begin{figure*}[]     
\centering                                     
\includegraphics[width=1.\textwidth]{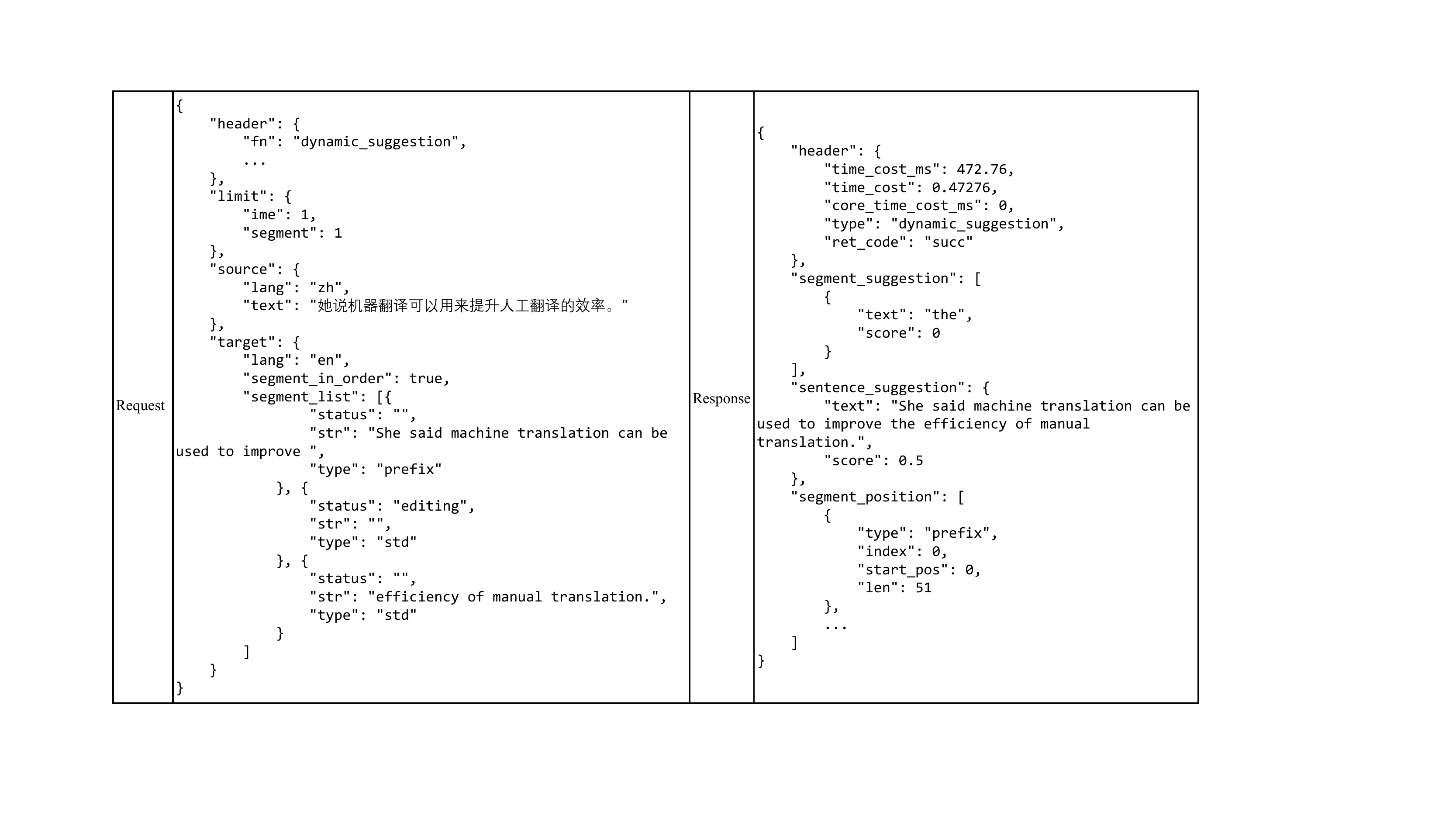}
\caption{Example of the sentence-level dynamic suggestion.}  
\label{fig:dynamic_suggestion_lcd}
\end{figure*}

One important API is the ``\texttt{dynamic\_suggestion}'' function, aiming to save the effort of human translators when correcting the translation generated by AI. This API is capable of providing autocompletion at both word-level and sentence-level, which will be introduced one by one.

First, the word-level autocompletion is able to complete the unfinished word input by human translators. As shown in Figure \ref{fig:dynamic_suggestion_ime}, when the a character sequence ``\texttt{th}'' is tagged by ``\texttt{editing}'' in the ``\texttt{segment\_list}'' field, TranSmart will return its autocompletion in the ``\texttt{ime\_suggestion}'' field in response. Second, the sentence-level suggestion can complete the whole translation based on the user-specified spans. As shown in Figure \ref{fig:dynamic_suggestion_lcd}, the span with the ``\texttt{prefix}'' type is promised to be the prefix of the re-generated translation in response, and spans with ``\texttt{std}'' are forced to be included in re-generated translation. The result of sentence-level autocompletion is in the ``\texttt{sentence\_suggestion}'' field in response.

\subsection{Memory-Aware Machine Translation}

\begin{figure*}[]     
\centering                                     
\includegraphics[width=1.\textwidth]{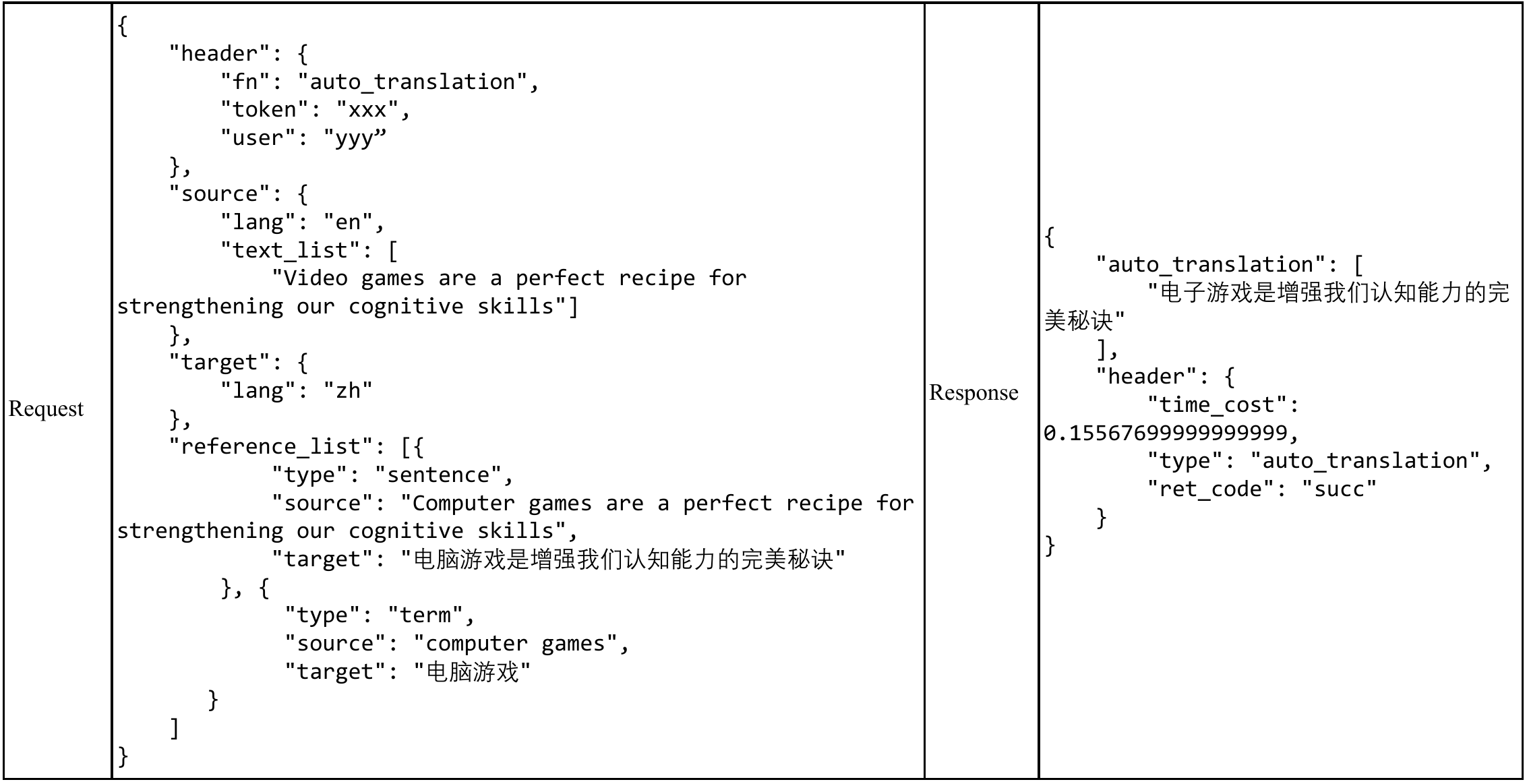}
\caption{Example of the automatic translation with translation memory.}  
\label{fig:auto_translation}
\end{figure*}

As shown in Figure \ref{fig:auto_translation}, when users provide the information of translation memory in the ``\texttt{reference\_list}'' field, TranSmart will be able to leverage those existing and relevant translations in history to improve the translation quality. The translation results are in the ``\texttt{auto\_translation}'' field in response. Note that users can also disable this feature by removing the ``\texttt{reference\_list}'' field, then TranSmart will translate only based on the source text, i.e., the setting of traditional automatic translation.


\subsection{Extended Features}

\begin{figure*}[]     
\centering                                     
\includegraphics[width=1.\textwidth]{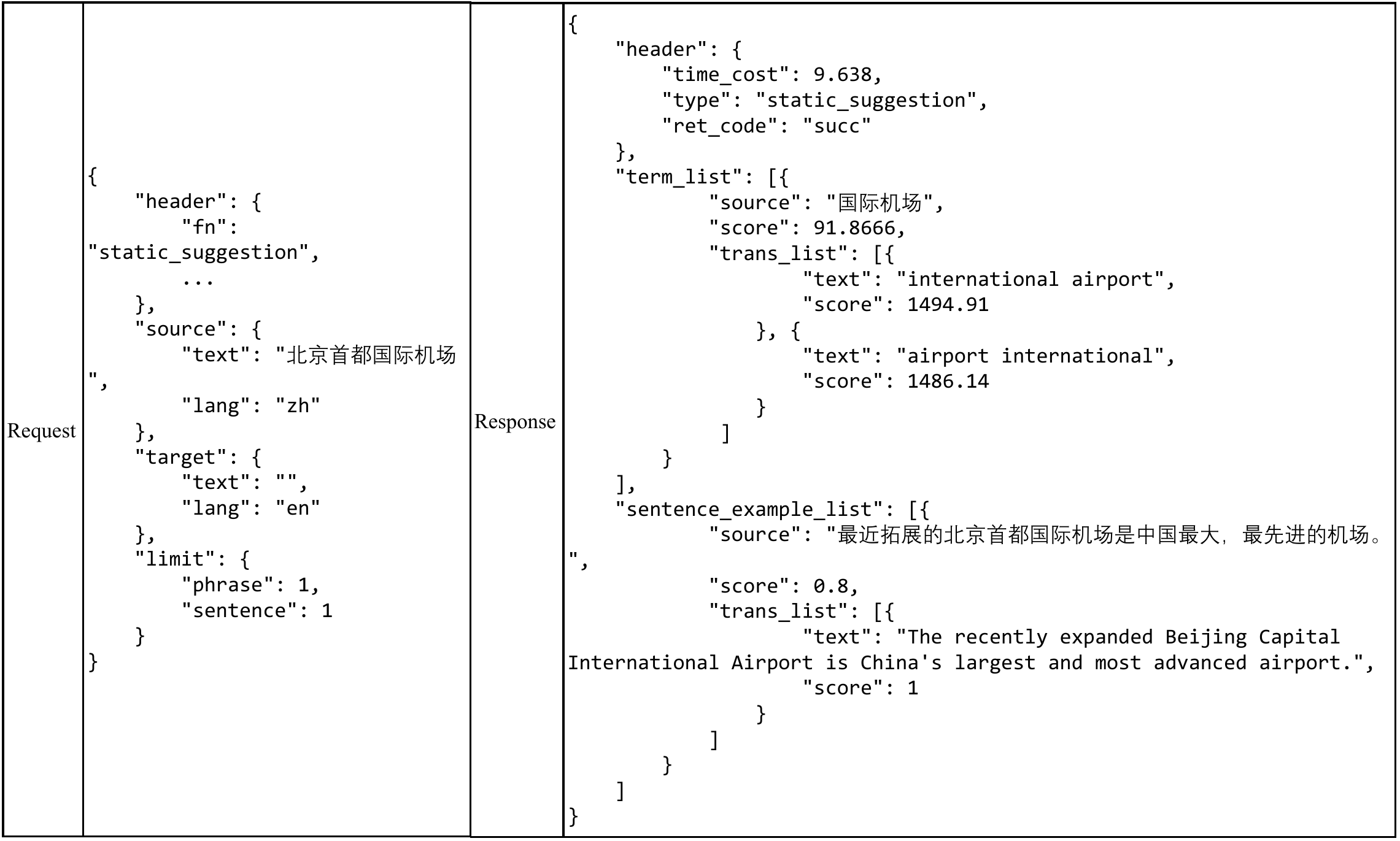}
\caption{Example of the static suggestion.}  
\label{fig:static_suggestion}
\end{figure*}

\begin{figure*}[]     
\centering                                     
\includegraphics[width=1.\textwidth]{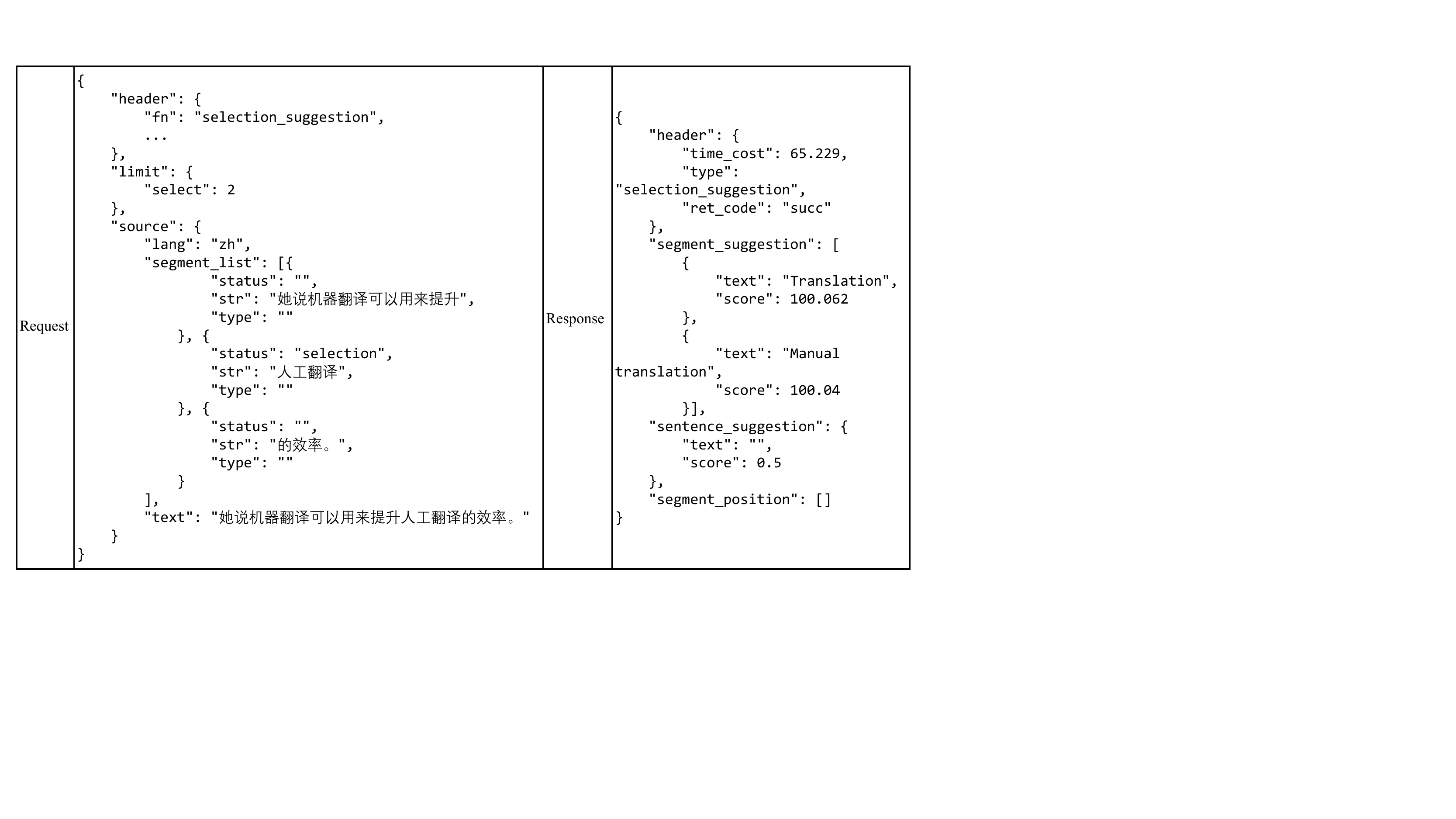}
\caption{Example of the selection suggestion.}  
\label{fig:selection_suggestion}
\end{figure*}

TranSmart also has two APIs that may be helpful for human translators. First, it provides a \texttt{static\_suggestion} function to retrieve relevant information from our pre-constructed terminology and bilingual example databases. As shown in Figure \ref{fig:static_suggestion}, the ``\texttt{static\_suggestion}'' function returns the most relevant terminology translations and bilingual examples in the ``\texttt{term\_list}'' and ``\texttt{sentence\_example\_list}'' field in response, respectively. Second, our ``\texttt{selection\_suggestion}'' is designed to translate specified spans in a source sentence. As shown in Figure \ref{fig:selection_suggestion}, only spans with ``\texttt{selection}'' status in the ``\texttt{segment\_list}'' field will be translated. The results of selection suggestions are placed in the ``\texttt{segment\_suggestion}'' field in response.


\section{System Evaluation}

\subsection{Generic Translation}

\paragraph{Datasets}
We conducted experiments on the  widely used WMT14 English$\Rightarrow$German 
(En$\Rightarrow$De) and English$\Rightarrow$French 
(En$\Rightarrow$Fr) datasets, which consist of about $4.5$M and $35.5$M sentence pairs, respectively.~\footnote{Note that although the datasets used in the experiments are preprocessed by the standard toolkit in Moses~\cite{koehn2007moses} (for a fair comparison between previous methods), in implementing the online system, we adopt TexSmart~\cite{zhang+:2020:texsmart} in data preprocessing (such as Chinese word segmentation) and postprocessing (e.g., restoring case information).} We applied BPE~\cite{sennrich+:2016} with 32K merge operations for both language pairs. The experimental results were reported in case-sensitive BLEU score~\cite{papineni2002bleu}.

\paragraph{Systems}
We validated our approach on a couple of representative NMT architectures:
\begin{itemize}
    \item \textsc{Lstm}~\cite{domhan2018much} that is implemented in the \textsc{Transformer} framework.
    \item \textsc{Transformer}~\cite{Vaswani:2017:NIPS} that is based solely on attention mechanisms.
    \item \textsc{DynamicConv}~\cite{wu2019pay} that is implemented with lightweight and dynamic convolutions, which can perform competitively to the best reported \textsc{Transformer} results.
\end{itemize}
We adopted the open-source toolkit Fairseq~\cite{ott:2019:naacl} to implement the above NMT models. 
We followed the settings in the original works to train the models.
In brief, we trained the \textsc{Lstm} model for 100K steps with 32K ($4096\times8$) tokens per batch.
For \textsc{Transformer}, we trained 100K and 300K steps with 32K tokens per batch for the \textsc{Base} and \textsc{Big} models respectively.
We trained the \textsc{DynamicConv} model for 30K steps with 459K ($3584\times128$) tokens per batch.
We selected the model with the best perplexity on the validation set as the final model.

\begin{table*}[t]
\setcounter{table}{2}
\centering
\begin{tabular}{c l ll ll}
\toprule
    \multirow{2}{*}{\bf System}  &   \multirow{2}{*}{\bf Architecture}  & \multicolumn{2}{c}{\bf WMT14 En$\Rightarrow$De}  &  \multicolumn{2}{c}{\bf WMT14 En$\Rightarrow$Fr}\\
    \cmidrule(lr){3-4} \cmidrule(lr){5-6}
        &   &   BLEU & $\bigtriangleup$    & BLEU & $\bigtriangleup$ \\
    \midrule
    \multicolumn{6}{c}{{\bf Existing NMT Systems}} \\
    \cite{domhan2018much}    &  \textsc{Lstm}    &    26.7   &   --  &   --  &   --\\
    \hdashline
    \multirow{2}{*}{\cite{Vaswani:2017:NIPS}} &   \textsc{Transformer-Base}    &    27.3  & --  &  38.1 &  -- \\
     &  \textsc{Transformer-Big} &    28.4  & --  &  41.0 &  -- \\
    \hdashline
    \cite{ott2018scaling} & ~~~~ + Large Batch &  29.3  &  --  &  43.2  &  --\\
    \hdashline
    \cite{wu2019pay} & \textsc{DynamicConv} & 29.7 & -- &  43.2 & --\\
    \midrule
    \multicolumn{6}{c}{{\bf Our NMT Systems}}   \\ 
    \hline
    \multirow{10}{*}{\em This work}
    &   \textsc{Lstm}              &    26.5   &   --  &   40.6  &   --\\
    &   ~~~ +  Data Rejuvenation   &    27.0$^\uparrow$       &    +0.5   &  41.1$^\uparrow$  &   +0.5\\ 
    \cline{2-6}
    &   \textsc{Transformer-Base}  &  27.5  & -- &  40.2  &   -- \\
    &   ~~~ +  Data Rejuvenation   &  28.3$^\Uparrow$  & +0.8  &  41.0$^\Uparrow$ & +0.8 \\ 
    \cdashline{2-6}
    &   \textsc{Transformer-Big}  &  28.4  & -- &   42.4 &   -- \\
    &   ~~~ +  Data Rejuvenation  &  29.2$^\Uparrow$  &  +0.8   & 43.0$^\uparrow$  &  +0.6 \\ 
    \cdashline{2-6}
    &   ~~~ + Large Batch  &   29.6     & -- &  43.5  &   -- \\
    &   ~~~~~~~~ +  Data Rejuvenation  &  30.3$^\Uparrow$  &  +0.7 & 44.0$^\uparrow$ & +0.5  \\
    \cline{2-6}
    &   \textsc{DynamicConv}      &   29.7             & --     & 43.3 &   -- \\
    &   ~~~ +  Data Rejuvenation  &   30.2$^\uparrow$  &  +0.5  & 43.9$^\uparrow$  &  +0.6 \\
    \bottomrule
  \end{tabular}
  \caption{Evaluation of translation performance across model architectures and language pairs. ``$\uparrow/\Uparrow$'':  indicate statistically significant improvement over the corresponding baseline $p < 0.05/0.01$ respectively.}  
  \label{tab:main}
\end{table*}

\paragraph{Results}
Table~\ref{tab:main} lists the results across model architectures and language pairs. 
Our \textsc{Transformer} models achieve better results than that reported in previous work~\cite{Vaswani:2017:NIPS}, especially on the large-scale En$\Rightarrow$Fr dataset (e.g., more than 1.0 BLEU points).
~\cite{ott2018scaling} showed that models of larger capacity benefit from training with large batches. Analogous to \textsc{DynamicConv}, we trained another \textsc{Transformer-Big} model with 459K tokens per batch (``+ Large Batch'' in Table~\ref{tab:main}) as a strong baseline.
We tested statistical significance with paired bootstrap resampling~\cite{Koehn2004:emnlp} using \texttt{compare-mt}\footnote{\url{https://github.com/neulab/compare-mt}}~\cite{neubig2019:naacl}.

Clearly, our data rejuvenation consistently and significantly improves translation performance in all cases, demonstrating the effectiveness and universality of the proposed data rejuvenation approach.
It's worth noting that our approach achieves significant improvements without introducing any additional data and model modification. It makes the approach robustly applicable to most existing NMT systems.

\subsection{Word Level Autocompletion}
\paragraph{Datasets}
We carry out experiments on four GWLAN tasks including bidirectional Chinese–English tasks and German–English tasks. The training set for two directional Chinese–English tasks consists of 1.25M bilingual sentence pairs from LDC corpora. As discussed in \S 3.2, the training data for GWLAN is extracted from 1.25M sentence pairs. The validation data for GWLAN is extracted from NIST02 and the test datasets for GWLAN are constructed from NIST05 and NIST06. For two German–English tasks, we use the WMT14 dataset standard preprocessed by Stanford \footnote{\url{https://nlp.stanford.edu/projects/nmt/}}. The validation and test sets for our tasks are based on newstest13 and newstest14 respectively. All the sample operations in the data construction process are based on the uniform distribution. For each dataset, the models are tuned and selected based on the validation set.

\paragraph{Systems} In the experiments, we evaluate and compare the performance of our proposed approach (\textsc{WPM}) and a few baselines, which are illustrated below:
\begin{itemize}
    \item \textsc{TransTable}: We train an alignment model \footnote{\url{https://github.com/clab/fast_align}} on the training set and build a word-level translation table. While testing, we can find the translations of all source words based on this table, and select out valid translations based on the human input. The word with highest frequency among all candidates is regarded as the prediction. This baseline is inspired by ~\cite{huang2015new,huang2018input}. 
    \item \textsc{Trans-PE}: We train a vanilla NMT model using the Transformer-base model. During the inference process, we use the context on the left hand side of human input as the model input, and return the most possible words based on the probability of valid words selected out by the human input. This baseline is inspired by ~\cite{langlais2000transtype, santy-etal-2019-inmt}.
    \item \textsc{Trans-NPE}: As another baseline, we also train an NMT model based on Transformer, but without position encoding on the target side. While testing, we use the averaged hidden vectors of all the target words outputted by the last decoder layer to predict the potential candidates.
\end{itemize}

\paragraph{Evaluation Metric}
To evaluate the performance of the well-trained models, we choose accuracy as the evaluation metric:
\begin{equation*}
    \mathrm{Acc} = \frac{N_{match}}{N_{all}},
\end{equation*}
where $N_{match}$ is the number of words that are correctly predicted  and $N_{all}$ is the number of testing examples.

\begin{table*}[t]
\resizebox{1\columnwidth}{!}{
\begin{tabular}{l||cc||cc||cc||cc@{}}
\toprule
     \multirow{2}{*}{Systems}     & \multicolumn{2}{c||}{Zh$\Rightarrow$En}                                & \multicolumn{2}{c||}{En$\Rightarrow$Zh}                                & \multicolumn{2}{c||}{De$\Rightarrow$En} & \multicolumn{2}{c}{En$\Rightarrow$De} \\ \cline{2-9}
      & \multicolumn{1}{c}{NIST05} & \multicolumn{1}{c||}{NIST06} & \multicolumn{1}{c}{NIST05} & \multicolumn{1}{c||}{NIST06} & NT13        & NT14       & NT13        & NT14       \\\hline\hline
 \textsc{TransTable} & 41.40                      & 39.78                      & 28.00                      & 26.99                      &     37.43        &       36.64     &       32.99      &    31.12        \\
 \textsc{Trans-PE}   & 34.51                      & 35.50                      & 32.23                      & 34.88                      &        34.45     &      33.02      &      31.51      &          30.65  \\
 \textsc{Trans-NPE}  & 35.97                      & 36.78                      & 34.31                      & 36.19                      &         36.69    &      36.01      &       33.25      &        31.30    \\
\hline
 \textsc{WPM} & \textbf{55.54}                      & \textbf{55.85}                      & \textbf{53.64}                      & \textbf{54.25}                             &       \textbf{57.84}     &    \textbf{56.75}   &    \textbf{56.91}     &      \textbf{52.68}      \\ \bottomrule
\end{tabular}
} \caption{\label{tab:all} The main results measured by word prediction precision for different systems on Chinese-English and German-English datasets.}
\end{table*}

\paragraph{Results}
Table \ref{tab:all} shows the main results of our method and three baselines on the test sets of Chinese-English and German-English datasets. 
The method \textsc{Trans-PE}, which assumes the human input is the next word of the given context, behaves poorly under the more general setting. As the results of \textsc{Trans-NPE} show, when we use the same model as \textsc{Trans-PE} and relax the constraint of position by removing the position encoding, the accuracy of the model improves. One interesting finding is that the \textsc{TransTable} method, which is only capable of leveraging the zero-context, achieves good results on the Chinese-English task when the target language is English. However, when the target language is Chinese, the performance of \textsc{TransTable} drops significantly.
It is clear from the results that our method \textsc{WPM} significantly outperforms the three baseline methods. 

\subsection{Sentence Level Autocompletion}
\paragraph{Datasets}
We conduct experiments on the  Zh$\Rightarrow$En, Fr$\Rightarrow$En, {De$\Rightarrow$En}, and  translation tasks. 
The Zh$\Rightarrow$En bilingual corpus includes news articles collected from several online news websites. After the standard preprocessing procedure as in \cite{koehn2007moses},
we obtain about 2 million bilingual sentences in total. Then we randomly select 2000 sentences as the development and test datasets, respectively, and leave other sentences as the training dataset. 
The Fr$\Rightarrow$En bilingual corpus is from JRC-Acquis datasets \cite{steinberger2006jrc} and it is preprocessed
following \cite{gu+:2018}. This dataset is a collection of the parallel legislative text of European Union 
Law applicable in the EU member states and thus it is a highly related corpus focusing on a specific domain.
{We also evaluate our methods on the WMT 2018 English-German news translation task, which is composed of Europarl and news commentary data\cite{dinu2019training}. We use the WMT \textit{newstest2013} and \textit{newstest2014} as the development and test set respectively.}
{In addition, in our experiments, we use the subword technique \cite{sennrich+:2016} to ensure that there are no unknown tokens in our input, even if a constraint word is not included in the training set. Therefore, if a human translator provides an unknown word as a constraint word, our system tokenizes it into several BPE tokens which are considered as several constraints accordingly.}
Since there are no real constraints available provided by human for all these datasets, we simulate this effect by randomly picking some words from the reference side as the constraints following~\cite{hokamp2017lexically,song-etal-2019-code}. 

\paragraph{Systems} 
As presented in section 3.3, we implement two methods: constrained decoding with ordered constraints denoted by \textsc{O-Gbs} and NMT with soft constraints denoted by \textsc{SC}. We compare both of our methods against three baselines: 
\begin{itemize}
    \item \textsc{Transformer}: It is the standard Transformer, which is an in-house implementation using Pytorch following \cite{klein2017opennmt}.
    \item \textsc{Gbs}: It is the Grid Beam Search in \cite{hokamp2017lexically} built on top of the in-house \textsc{Transformer}. 
    \item  \textsc{Dba}: It is the lexically constrained decoding with Dynamic Beam Allocation (an improved approach for \cite{hokamp2017lexically}) in \cite{post2018fast} .
\end{itemize}
To ensure the translation quality, we set the default beam size for \textsc{Gbs} and \textsc{Dba} as suggested by \cite{hokamp2017lexically} and \cite{post2018fast}. 
{The hyper-parameters for all the systems are following those of \textsc{Transformer} base model\cite{Vaswani:2017:NeurIPS}}. We train all the models with Adam optimization algorithm \cite{kingma2015adam} and tune the number of iteration based on the performance of the development set. 


\begin{table}[t]
    \begin{center}
    \begin{tabular}{l||l||l||l||l} 
    \hline
    {Systems} & Fr$\Rightarrow$En  & {De$\Rightarrow$En} & Zh$\Rightarrow$En  & Runtime \\ \hline\hline
    \textsc{Transformer} & 67.13 & {26.46} & 36.22  & 0.256\\
    \quad + \textsc{Gbs} & 70.99$^{\dagger}$ & {30.67}$^{\dagger}$ & 42.17$^{\dagger}$  & 5.692\\
    \quad + \textsc{Dba}& 67.35$^{\dagger}$ & {29.04}$^{\dagger}$  & 41.23$^{\dagger}$  & 1.531\\
    \hline
    \quad + \textsc{O-Gbs} & 70.99$^{\dagger}$ & {30.67}$^{\dagger}$ & 42.17$^{\dagger}$  & 1.959 \\
    \quad + \textsc{SC} & 71.10$^{\dagger}$ & {30.25$^{\dagger}$} & 42.96$^{\dagger}$  & 0.262\\
    \hline
    \end{tabular}
    \end{center}
    \caption{\label{table:zh-sc2} BLEU and runtime comparisons on perfect constraints for the Zh$\Rightarrow$En, Fr$\Rightarrow$En, and {De$\Rightarrow$En} tasks.
    The runtime is measured by the time consumed by decoding one sentence on the Zh$\Rightarrow$En task. {$\dagger$ indicates the comparison over \textsc{Transformer} is significant with p \textless \ 0.05.}}
    \end{table}




\paragraph{Results}
 As shown in Table \ref {table:zh-sc2}, although the \textsc{Dba} method would reduce the computational overhead compared to \textsc{Gbs}, its translation quality slightly sacrifices accordingly, which is similar to the finding in \cite{post2018fast}. Thanks to the ordered constraints, the \textsc{O-Gbs} is faster than \textsc{Gbs} in running speed and delivers the same performance as \textsc{Gbs}. The advantage of these three lexically decoding algorithm is that they do not need to retrain a translation model. 
 Moreover, \textsc{SC} performs better than \textsc{Gbs} on the Fr$\Rightarrow$En and Zh$\Rightarrow$En tasks and outperforms \textsc{Dba} on all the three tasks, in terms of translation quality. This observation indicates that our model \textsc{SC} is able to learn how to integrate the lexical constraints correctly. Moreover, there is an additional advantage that the decoding speed of our model is comparable with that of the \textsc{Transformer} model and is much faster than lexically constrained decoding.

\subsection{Translation Memory}
Since there is no translation history provided by human translators, we simply use the training set as the memory for simulation. For each sentence, we retrieve 100 translation pairs from the training set by using Apache Lucene \cite{bialecki2012apache}. We score the source side of each retrieved pair against the source sentence with fuzzy matching score and select top $N=5$ translation sentence pairs as a translation memory for the sentence to be translated, following \cite{gu+:2018,zhang+:2018,he+:2019:word}. Sentences from the target side in the translation memory are used to form a graph, with each word represented as a node and the connection between adjacent words in a sentence represented as an undirected edge. 

\paragraph{Datasets}
Following the previous works incorporating TM into NMT models, we use the JRC-Acquis corpus for training and evaluating our proposed model. The JRC-Acquis corpus is a collection of parallel legislative text of European union Law applicable in the EU member states. The highly related text in the corpus is suitable for us to make evaluations. To fully explore the effectiveness of our proposed model, we conduct translation experiments on three language pair bidirectionally, namely, en-fr, en-es, and en-de.
We manage to obtain preprocessed datasets from \cite{gu+:2018}. For each language pair, we randomly select 3000 samples to form a development and a test set respectively. The rest of the pairs are used as the training set. Sentences longer than 80 and 100 are removed from the training and development/test set. The technique of Byte-pair Encoding \cite{sennrich+:2016} is applied and the vocabulary size is set to be 20K for all the experiments. 

\paragraph{Systems}
The proposed graph based TM model is built on transformer \cite{Vaswani:2017:NeurIPS}, and it is denoted by {\bf G-TFM}. We compare the proposed model against the following baselines:\footnote{We notice that there are some recent advances on NMT with translation memory~\cite{xu-etal-2020-boosting,khandelwal2020nearest,he2021fast,cai2021neural} after the proposed G-TFM was implemented in TranSmart. We consider to update our TM model in future.}
\begin{itemize}
    \item {\bf TFM}: It is a natural baseline as the proposed model is directly built upon the Transformer architecture.
    \item {\bf P-RNN}: It is an in-house implementation of \cite{zhang+:2018} on top of RNN-search. 
    \item {\bf P-TFM}: It similar to P-RNN but it is on top of Transformer rather than RNN-search as \cite{zhang+:2018}. 
    \item {\bf SEG-TFM}: It implements the idea of \cite{gu+:2018} on top of Transformer. Due to the architecture divergence between RNN-based NMT and Transformer, it only differs from the RNN-based counterpart in that two quantities $c_t$ and $z_t$ in \cite{gu+:2018} 
are replaced by the hidden units obtained from the multi-head attention over the encoding units and the decoding hidden state units before the softmax operator.
    \item {\bf SEQ-TFM}: It sequentially encodes all target sentences in a TM as one of the baseline models. Specifically, each target sentence in TM goes through a multi-head mechanism and an immediate residual connection plus layer normalization in $l_{th}$ layer. The derived representations for these sentences are then concatenated to form the representation of the translation memory, which can be utilized flexibly in $l_{th}$ decoding layer. 
\end{itemize}

For training all systems, we maintain the same hyper-parameters for fair comparison. Besides, we adopt the same training algorithm to learn the models as follows. 
We use a customized leaning rate decay paradigm following Tensor2Tensor\cite{vaswani2018tensor2tensor} package. The learning rate increases linearly on early stages for a certain number of steps, known as warm-up steps, and decay exponentially later on. We set the warm-up step to be 5 epochs and we early stop the model after training 20 epochs, typically the time when the development performance varies insignificantly.
Furthermore, since there is a hyperparameter in the system P-TFM of \cite{zhang+:2018} which is sensitive to the specific translation task, 
we tune it carefully on the development set for
all translation tasks. Its optimized value is 0.7 for es and de tasks while it is 0.8 for fr task.~\footnote{We run all 6 tasks with hyperparameters among [0.5, 1.5] with scale of 0.1, and manually pick the optimized value according to its performance on the development set.}

\paragraph{Results}

\begin{table*}[t!]
\centering
\begin{tabular}{c|cc|ccccc}
  \toprule
  & RNN & P-RNN & TFM & P-TFM & SEG-TFM & SEQ-TFM  & G-TFM   \\
\hline
\hline
  Dev & 57.74 & 60.87 & 62.78 & 63.97 & 63.16  & 64.81 & \textbf{66.37} \\
  Test & 58.06 & 61.52 & 62.68 & 64.30 & 62.94  & 65.16 & \textbf{66.21}   \\
  \bottomrule
\end{tabular}
\caption{Translation accuracy in terms of BLEU on the es-en task.}
\label{table:res:es-en}
\end{table*}

Table~\ref{table:res:es-en} shows the experiment results of all the systems on the es-en task in terms of BLEU.
Several observations can be made from the results. First, the baseline TFM achieves substantial gains over RNN and 
even outperforms P-RNN by around 1 BLEU point on the test set.  
Compared with the strongest baseline P-TFM, 
the proposed SEQ-TFM and G-TFM are able to obtain some gains up to 1.9 BLEU points on the test set. 
This result verifies that our compact representation of TM is able to guide the decoding of the state-of-the-art model. 

\begin{table*}[t!]
\centering
\begin{tabular}{c|cccc}
  \toprule
   & TFM & SEG-TFM & SEQ-TFM  & G-TFM  \\
\hline
\hline
  Train (s) & 4579 & 44238 & 21920 & 8692  \\
  Test (s) & 0.20 & 2.68 & 1.25 & 0.36 \\ 
  Words (\#) & 68.28 & 374.52 & 214.97 & 129.18  \\
  \cmidrule(lr){1-5}
  BLEU & 62.68 & 62.94 & 65.16 & \bf{66.21}  \\
  \bottomrule
\end{tabular}
\caption{Running time and memory. Training time reports the time in seconds for training one epoch on average, and testing time reports the time in seconds for translating one sentence on average. Words (\#) denotes the number of words encoded in the neural models on average.}
\label{table:time}
\end{table*}

Second, it is observed that SEG-TFM is only comparable to TFM on this task, although its RNN based counterpart 
brought significant gains as reported in \cite{gu+:2018}. This fact shows that the transformer architecture 
may need a sophisticated way to well define a key-value memory for TM encoding, which can be significantly different from that on RNN architecture. This is beyond the scope of this paper. Fortunately, this paper provides an easy yet effective approach to encode a TM, 
i.e. G-TFM, which does not rely on a context-based key-value memory. 

Since the retrieval time can be neglected compared with the decoding time as found in \cite{zhang+:2018}, we thereby eliminate the retrieval time and directly compare running time for neural models as shown in 
Table \ref{table:time}. From this table, we observe that the proposed graph based model G-TFM saves significant running time compared with SEG-TFM and SEQ-TFM while achieving better translation performance.

Table \ref{table:time} 
depicts the total number of source and target words encoded by the corresponding model for each test sentence on average. It's observed that SEG-TFM needs to encode approximately 3 times and SEQ-TFM encodes approximately 2 times the number of words of our proposed model, G-TFM. There's no wonder that TFM takes the fewest words to encode because no extra TM is included. These statistics indicate that under the scenario of incorporating TM in NMT, our m1odel requires the least memory.

\begin{table*}[t!]
\centering
\begin{tabular}{cc|cc||cc||cc}
  \toprule
  & BLEU & en-fr & fr-en & en-de & de-en & en-es\\
\hline
\hline
   Dev & TFM & 66.33 & 65.95 & 53.32 & 58.54 & 60.43 \\
   & P-TFM & 68.90 & 68.61 &		55.54 & 60.10 & 61.50 \\
  	  & G-TFM & \bf{69.69} & \bf{70.65} & \bf{57.43} & \bf{61.85} & \bf{62.50}  \\ 
      \cmidrule(lr){1-7}
   Test & TFM & 66.36 & 66.96 & 53.29 & 58.86 & 60.52 \\
   & P-TFM & 68.73 & 68.70 & 55.14 & 60.26 & 61.56\\
  & G-TFM & \bf{69.59} & \bf{70.87} & \bf{56.88} & \bf{61.72} & \bf{62.76} \\
\bottomrule
\end{tabular}
\caption{Translation Results on both development and test sets across other 5 translation tasks. \label{main results}}
\end{table*}

We pick stronger baselines from the es-en task, i.e. TFM and P-TFM, and compare them with the proposed G-TFM model on other 5 translation tasks. Table \ref{main results} summarizes their results on both the development and test sets. 
From this table, we can see that on the test set, G-TFM steadily outperforms TFM by up to 3 BLEU points across all these 5 tasks, 
In addition, contrast to P-TFM, G-TFM demonstrates better performance by exceeding at least 1 BLEU point across all these tasks except the en-fr task. 
These results are consistent with the results on es-en task and further validates the effectiveness of integrating graph-based translation memory into the Transformer model. 

\section{Conclusion}
In this technical report we have presented TranSmart, a practical interactive machine translation (IMT) system. Unlike the conventional IMT systems with the strict manner from left to right, TranSmart conducts interaction between a user and the machine in a flexible manner, and it particularly contains a translation memory technique to avoid similar mistakes occurring during translation process.
We have introduced the main functions of TranSmart and key methods for implementing the features.
Some instructions about how to use the TranSmart through online APIs have been described.
We have also reported some evaluation results on major modules of TranSmart.

\bibliographystyle{unsrt}  



\bibliography{main.bib}

\begin{thebibliography}{10}

\bibitem{Vaswani:2017:NeurIPS}
Ashish Vaswani, Noam Shazeer, Niki Parmar, Jakob Uszkoreit, Llion Jones,
  Aidan~N Gomez, {\L}ukasz Kaiser, and Illia Polosukhin.
\newblock Attention is all you need.
\newblock In {\em NeurIPS}, 2017.

\bibitem{gehring2017convolutional}
Jonas Gehring, Michael Auli, David Grangier, Denis Yarats, and Yann~N Dauphin.
\newblock Convolutional sequence to sequence learning.
\newblock In {\em International Conference on Machine Learning}, pages
  1243--1252. PMLR, 2017.

\bibitem{sutskever2014sequence}
Ilya Sutskever, Oriol Vinyals, and Quoc~V Le.
\newblock Sequence to sequence learning with neural networks.
\newblock {\em arXiv preprint arXiv:1409.3215}, 2014.

\bibitem{bahdanau2014neural}
Dzmitry Bahdanau, Kyunghyun Cho, and Yoshua Bengio.
\newblock Neural machine translation by jointly learning to align and
  translate.
\newblock {\em arXiv preprint arXiv:1409.0473}, 2014.

\bibitem{ondrej2017findings}
Bojar Ondrej, Rajen Chatterjee, Federmann Christian, Graham Yvette, Haddow
  Barry, Huck Matthias, Koehn Philipp, Liu Qun, Logacheva Varvara, Monz
  Christof, et~al.
\newblock Findings of the 2017 conference on machine translation (wmt17).
\newblock In {\em Second Conference onMachine Translation}, pages 169--214. The
  Association for Computational Linguistics, 2017.

\bibitem{barrault2019findings}
Lo{\"\i}c Barrault, Ond{\v{r}}ej Bojar, Marta~R Costa-Jussa, Christian
  Federmann, Mark Fishel, Yvette Graham, Barry Haddow, Matthias Huck, Philipp
  Koehn, Shervin Malmasi, et~al.
\newblock Findings of the 2019 conference on machine translation (wmt19).
\newblock In {\em Proceedings of the Fourth Conference on Machine Translation
  (Volume 2: Shared Task Papers, Day 1)}, pages 1--61, 2019.

\bibitem{wu2020tencent}
Shuangzhi Wu, Xing Wang, Longyue Wang, Fangxu Liu, Jun Xie, Zhaopeng Tu,
  Shuming Shi, and Mu~Li.
\newblock Tencent neural machine translation systems for the wmt20 news
  translation task.
\newblock In {\em Proceedings of the Fifth Conference on Machine Translation},
  pages 313--319, 2020.

\bibitem{peris2017interactive}
{\'A}lvaro Peris, Miguel Domingo, and Francisco Casacuberta.
\newblock Interactive neural machine translation.
\newblock {\em Computer Speech \& Language}, 45:201--220, 2017.

\bibitem{weng2019correct}
Rongxiang Weng, Hao Zhou, Shujian Huang, Lei Li, Yifan Xia, and Jiajun Chen.
\newblock Correct-and-memorize: Learning to translate from interactive
  revisions.
\newblock {\em arXiv preprint arXiv:1907.03468}, 2019.

\bibitem{zhao+:2020:balancing}
Tianxiang Zhao, Lemao Liu, Guoping Huang, Zhaopeng Tu, Huayang Li, Yingling
  Liu, Guiquan Liu, and Shuming Shi.
\newblock Balancing quality and human involvement: An effective approach to
  interactive neural machine translation.
\newblock In {\em AAAI}, 2020.

\bibitem{plitt2010productivity}
Mirko Plitt and Fran{\c{c}}ois Masselot.
\newblock A productivity test of statistical machine translation post-editing
  in a typical localisation context.
\newblock {\em The Prague bulletin of mathematical linguistics}, 93(1):7--16,
  2010.

\bibitem{green2014predictive}
Spence Green, Jason Chuang, Jeffrey Heer, and Christopher~D Manning.
\newblock Predictive translation memory: A mixed-initiative system for human
  language translation.
\newblock In {\em Proceedings of the 27th annual ACM symposium on User
  interface software and technology}, pages 177--187, 2014.

\bibitem{knowles2016neural}
Rebecca Knowles and Philipp Koehn.
\newblock Neural interactive translation prediction.
\newblock In {\em Proceedings of the Association for Machine Translation in the
  Americas}, pages 107--120, 2016.

\bibitem{grangier+auli:2018}
David Grangier and Michael Auli.
\newblock {Q}uick{E}dit: Editing text {\&} translations by crossing words out.
\newblock In {\em Proceedings of the 2018 Conference of the North {A}merican
  Chapter of the Association for Computational Linguistics: Human Language
  Technologies}, pages 272--282, New Orleans, Louisiana, June 2018. Association
  for Computational Linguistics.

\bibitem{wang+:2020:touch}
Qian Wang, Jiajun Zhang, Lemao Liu, Guoping Huang, and Chengqing Zong.
\newblock Touch editing: {A} flexible one-time interaction approach for
  translation.
\newblock In {\em Proceedings of AACL-IJCNLP}, 2020.

\bibitem{foster1997target}
George Foster, Pierre Isabelle, and Pierre Plamondon.
\newblock Target-text mediated interactive machine translation.
\newblock {\em Machine Translation}, 12(1):175--194, 1997.

\bibitem{langlais2000transtype}
Philippe Langlais, George Foster, and Guy Lapalme.
\newblock Transtype: a computer-aided translation typing system.
\newblock In {\em ANLP-NAACL 2000 Workshop: Embedded Machine Translation
  Systems}, 2000.

\bibitem{alabau2014casmacat}
Vicent Alabau, Christian Buck, Michael Carl, Francisco Casacuberta, Mercedes
  Garc{\'\i}a-Mart{\'\i}nez, Ulrich Germann, Jes{\'u}s Gonz{\'a}lez-Rubio,
  Robin Hill, Philipp Koehn, Luis~A Leiva, et~al.
\newblock Casmacat: A computer-assisted translation workbench.
\newblock In {\em Proceedings of the Demonstrations at the 14th Conference of
  the European Chapter of the Association for Computational Linguistics}, 2014.

\bibitem{koehn-etal-2003-statistical}
Philipp Koehn, Franz~J. Och, and Daniel Marcu.
\newblock Statistical phrase-based translation.
\newblock In {\em Proceedings of the 2003 Human Language Technology Conference
  of the North {A}merican Chapter of the Association for Computational
  Linguistics}, pages 127--133, 2003.

\bibitem{chiang2005hierarchical}
David Chiang.
\newblock A hierarchical phrase-based model for statistical machine
  translation.
\newblock In {\em Proceedings of the 43rd annual meeting of the association for
  computational linguistics (acl’05)}, pages 263--270, 2005.

\bibitem{koehn2009statistical}
Philipp Koehn.
\newblock {\em Statistical machine translation}.
\newblock Cambridge University Press, 2009.

\bibitem{chen2000new}
Zheng Chen and Kai-Fu Lee.
\newblock A new statistical approach to chinese pinyin input.
\newblock In {\em Proceedings of the 38th Annual Meeting of the Association for
  Computational Linguistics}, pages 241--247, 2000.

\bibitem{wang2014systematic}
Longyue Wang, Derek~F Wong, Lidia~S Chao, Yi~Lu, and Junwen Xing.
\newblock A systematic comparison of data selection criteria for smt domain
  adaptation.
\newblock {\em The Scientific World Journal}, 2014, 2014.

\bibitem{wang2014combining}
Longyue Wang, Yi~Lu, Derek~F Wong, Lidia~S Chao, Yiming Wang, and Francisco
  Oliveira.
\newblock Combining domain adaptation approaches for medical text translation.
\newblock In {\em Proceedings of the Ninth Workshop on Statistical Machine
  Translation}, pages 254--259, 2014.

\bibitem{Ott:2018:WMT}
Myle Ott, Sergey Edunov, David Grangier, and Michael Auli.
\newblock Scaling neural machine translation.
\newblock In {\em WMT}, 2018.

\bibitem{kingma2015adam}
Diederik~P. Kingma and Jimmy Ba.
\newblock Adam: A method for stochastic optimization.
\newblock In {\em ICLR}, 2015.

\bibitem{Jiao:2020:EMNLP}
Wenxiang Jiao, Xing Wang, Shilin He, Irwin King, Michael Lyu, and Zhaopeng Tu.
\newblock {D}ata {R}ejuvenation: {E}xploiting {I}nactive {T}raining {E}xamples
  for {N}eural {M}achine {T}ranslation.
\newblock In {\em EMNLP}, 2020.

\bibitem{Jiao:2021:ACL}
Wenxiang Jiao, Xing Wang, Zhaopeng Tu, Shuming Shi, Michael Lyu, and Irwin
  King.
\newblock Self-training sampling with monolingual data uncertainty for neural
  machine translation.
\newblock In {\em ACL}, 2021.

\bibitem{huang2015new}
Guoping Huang, Jiajun Zhang, Yu~Zhou, and Chengqing Zong.
\newblock A new input method for human translators: integrating machine
  translation effectively and imperceptibly.
\newblock In {\em IJCAI}, 2015.

\bibitem{santy-etal-2019-inmt}
Sebastin Santy, Sandipan Dandapat, Monojit Choudhury, and Kalika Bali.
\newblock {INMT}: Interactive neural machine translation prediction.
\newblock In {\em EMNLP-IJCNLP: System Demonstrations}, November 2019.

\bibitem{vasconcellos1985spanam}
Muriel Vasconcellos and Marjorie Le{\'o}n.
\newblock Spanam and engspan: machine translation at the pan american health
  organization.
\newblock {\em Computational Linguistics}, 11(2-3):122--136, 1985.

\bibitem{li2021gwlan}
Huayang Li, Lemao Liu, Guoping Huang, and Shuming Shi.
\newblock Gwlan: General word-level autocompletion for computer-aided
  translation.
\newblock In {\em Proceedings of the 59th Annual Meeting of the Association for
  Computational Linguistics}, 2021.

\bibitem{cheng2016primt}
Shanbo Cheng, Shujian Huang, Huadong Chen, Xinyu Dai, and Jiajun Chen.
\newblock Primt: A pick-revise framework for interactive machine translation.
\newblock In {\em Proceedings of the 2016 Conference of the North American
  Chapter of the Association for Computational Linguistics: Human Language
  Technologies}, pages 1240--1249, 2016.

\bibitem{wuebker2016models}
Joern Wuebker, Spence Green, John DeNero, Sa{\v{s}}a Hasan, and Minh-Thang
  Luong.
\newblock Models and inference for prefix-constrained machine translation.
\newblock In {\em Proceedings of ACL}, 2016.

\bibitem{hokamp2017lexically}
Chris Hokamp and Qun Liu.
\newblock Lexically constrained decoding for sequence generation using grid
  beam search.
\newblock In {\em Proceedings of the 55th Annual Meeting of the Association for
  Computational Linguistics}, pages 1535--1546, Vancouver, Canada, July 2017.
  Association for Computational Linguistics.

\bibitem{post2018fast}
Matt Post and David Vilar.
\newblock Fast lexically constrained decoding with dynamic beam allocation for
  neural machine translation.
\newblock In {\em Proceedings of the 2018 Conference of the North {A}merican
  Chapter of the Association for Computational Linguistics: Human Language
  Technologies}, pages 1314--1324, New Orleans, Louisiana, June 2018.
  Association for Computational Linguistics.

\bibitem{li+:2020:neural}
Huayang Li, Guoping Huang, Deng Cai, and Lemao Liu.
\newblock Neural machine translation with noisy lexical constraints.
\newblock {\em IEEE/ACM Transactions on Audio, Speech, and Language
  Processing}, 28:1864--1874, 2020.

\bibitem{chu2017empirical}
Chenhui Chu, Raj Dabre, and Sadao Kurohashi.
\newblock An empirical comparison of domain adaptation methods for neural
  machine translation.
\newblock In {\em Proceedings of the 55th Annual Meeting of the Association for
  Computational Linguistics (Volume 2: Short Papers)}, pages 385--391, 2017.

\bibitem{wang2017instance}
Rui Wang, Masao Utiyama, Lemao Liu, Kehai Chen, and Eiichiro Sumita.
\newblock Instance weighting for neural machine translation domain adaptation.
\newblock In {\em Proceedings of the 2017 Conference on Empirical Methods in
  Natural Language Processing}, pages 1482--1488, 2017.

\bibitem{chu-wang-2018-survey}
Chenhui Chu and Rui Wang.
\newblock A survey of domain adaptation for neural machine translation.
\newblock In {\em Proceedings of the 27th International Conference on
  Computational Linguistics}, pages 1304--1319, Santa Fe, New Mexico, USA,
  August 2018. Association for Computational Linguistics.

\bibitem{liu+:2012:locally}
Lemao Liu, Hailong Cao, Taro Watanabe, Tiejun Zhao, Mo~Yu, and Conghui Zhu.
\newblock Locally training the log-linear model for {SMT}.
\newblock In {\em EMNLP}, July 2012.

\bibitem{li2016one}
Xiaoqing Li, Jiajun Zhang, and Chengqing Zong.
\newblock One sentence one model for neural machine translation.
\newblock {\em arXiv preprint arXiv:1609.06490}, 2016.

\bibitem{farajian2017multi}
M~Amin Farajian, Marco Turchi, Matteo Negri, and Marcello Federico.
\newblock Multi-domain neural machine translation through unsupervised
  adaptation.
\newblock In {\em Proceedings of the Second Conference on Machine Translation},
  pages 127--137, 2017.

\bibitem{gu+:2018}
Jiatao Gu, Yong Wang, Kyunghyun Cho, and Victor O.~K. Li.
\newblock Search engine guided neural machine translation.
\newblock In {\em Proceedings of the Thirty-Second {AAAI} Conference on
  Artificial Intelligence, New Orleans, Louisiana, USA, February 2-7, 2018},
  2018.

\bibitem{zhang+:2018}
Jingyi Zhang, Masao Utiyama, Eiichiro Sumita, Graham Neubig, and Satoshi
  Nakamura.
\newblock Guiding neural machine translation with retrieved translation pieces.
\newblock In {\em Meeting of the North American Chapter of the Association for
  Computational Linguistics (NAACL)}, June 2018.

\bibitem{he+:2019:word}
Qiuxiang He, Guoping Huang, Lemao Liu, and Li~Li.
\newblock Word position aware translation memory for neural machine
  translation.
\newblock In {\em {NLPCC}}, 2019.

\bibitem{xia+:2019:graph}
Mengzhou Xia, Guoping Huang, Lemao Liu, and Shuming Shi.
\newblock Graph based translation memory for neural machine translation.
\newblock In {\em The Thirty-Third {AAAI} Conference on Artificial
  Intelligence}, pages 7297--7304, 2019.

\bibitem{koehn:2009:book}
Philipp Koehn.
\newblock {\em Statistical machine translation}.
\newblock Cambridge University Press, 2009.

\bibitem{dyer2010cdec}
Chris Dyer, Adam Lopez, Juri Ganitkevitch, Jonathan Weese, Ferhan T{\"u}re,
  Phil Blunsom, Hendra Setiawan, Vladimir Eidelman, and Philip Resnik.
\newblock cdec: A decoder, alignment, and learning framework for finite-state
  and context-free translation models.
\newblock In {\em Proceedings of the ACL 2010 System Demonstrations}, pages
  7--12, 2010.

\bibitem{mangu+:1999}
Lidia Mangu, Eric Brill, and Andreas Stolcke.
\newblock Finding consensus among words: Lattice-based word error minimization.
\newblock In {\em Sixth European Conference on Speech Communication and
  Technology}, 1999.

\bibitem{mangu+:2000}
Lidia Mangu, Eric Brill, and Andreas Stolcke.
\newblock Finding consensus in speech recognition: word error minimization and
  other applications of confusion networks.
\newblock {\em Computer Speech \& Language}, 14(4):373--400, 2000.

\bibitem{velivckovic+:2018}
Petar Veli{\v{c}}kovi{\'c}, Guillem Cucurull, Arantxa Casanova, Adriana Romero,
  Pietro Li{\`o}, and Yoshua Bengio.
\newblock Graph attention networks.
\newblock In {\em Proceedings of ICLR}, 2018.

\bibitem{liu+:2016:neural}
Lemao Liu, Masao Utiyama, Andrew~M. Finch, and Eiichiro Sumita.
\newblock Neural machine translation with supervised attention.
\newblock In {\em {COLING}}, pages 3093--3102, 2016.

\bibitem{li+:2018:target}
Xintong Li, Lemao Liu, Zhaopeng Tu, Shuming Shi, and Max Meng.
\newblock Target foresight based attention for neural machine translation.
\newblock In {\em NAACL-HLT}, pages 1380--1390, 2018.

\bibitem{chen+:2020:accurate}
Yun Chen, Yang Liu, Guanhua Chen, Xin Jiang, and Qun Liu.
\newblock Accurate word alignment induction from neural machine translation.
\newblock In {\em EMNLP}, November 2020.

\bibitem{li+:2019:on}
Xintong Li, Guanlin Li, Lemao Liu, Max Meng, and Shuming Shi.
\newblock On the word alignment from neural machine translation.
\newblock In {\em ACL}, pages 1293--1303, 2019.

\bibitem{ding2019saliency}
Ding Shuoyang, Xu~Hainan, and Koehn Philipp.
\newblock Saliency-driven word alignment interpretation for neural machine
  translation.
\newblock In {\em WMT 2019}, page~1.

\bibitem{li+:2020:evaluating}
Jierui Li, Lemao Liu, Huayang Li, Guanlin Li, Guoping Huang, and Shuming Shi.
\newblock Evaluating explanation methods for neural machine translation.
\newblock In {\em ACL}, pages 365--375, July 2020.

\bibitem{chen2020maskalign}
Chi Chen, Maosong Sun, and Yang Liu.
\newblock Mask-align: Self-supervised neural word alignment.
\newblock In {\em arXiv}, 2020.

\bibitem{och2003systematic}
Franz~Josef Och and Hermann Ney.
\newblock A systematic comparison of various statistical alignment models.
\newblock {\em Computational linguistics}, 29(1):19--51, 2003.

\bibitem{dyer2013simple}
Chris Dyer, Victor Chahuneau, and Noah~A Smith.
\newblock A simple, fast, and effective reparameterization of ibm model 2.
\newblock In {\em Proceedings of the 2013 Conference of the North American
  Chapter of the Association for Computational Linguistics: Human Language
  Technologies}, pages 644--648, 2013.

\bibitem{vogel1996hmm}
Stephan Vogel, Hermann Ney, and Christoph Tillmann.
\newblock Hmm-based word alignment in statistical translation.
\newblock In {\em COLING}, 1996.

\bibitem{longyue2019discourse}
Longyue Wang.
\newblock {\em Discourse-aware neural machine translation}.
\newblock PhD thesis, Dublin City University, 2019.

\bibitem{wang2017exploiting}
Longyue Wang, Zhaopeng Tu, Andy Way, and Qun Liu.
\newblock Exploiting cross-sentence context for neural machine translation.
\newblock In {\em Proceedings of the 2017 Conference on Empirical Methods in
  Natural Language Processing}, pages 2826--2831, 2017.

\bibitem{wang2016automatic}
Longyue Wang, Xiaojun Zhang, Zhaopeng Tu, Andy Way, and Qun Liu.
\newblock Automatic construction of discourse corpora for dialogue translation.
\newblock In {\em Proceedings of the Tenth International Conference on Language
  Resources and Evaluation (LREC'16)}, pages 2748--2754, 2016.

\bibitem{wang2020tencent}
Longyue Wang, Zhaopeng Tu, Xing Wang, Li~Ding, Liang Ding, and Shuming Shi.
\newblock Tencent ai lab machine translation systems for wmt20 chat translation
  task.
\newblock In {\em Proceedings of the Fifth Conference on Machine Translation},
  pages 483--491, 2020.

\bibitem{wang2016novel}
Longyue Wang, Zhaopeng Tu, Xiaojun Zhang, Hang Li, Andy Way, and Qun Liu.
\newblock A novel approach to dropped pronoun translation.
\newblock In {\em Proceedings of NAACL-HLT}, pages 983--993, 2016.

\bibitem{wang2018translating}
Longyue Wang, Zhaopeng Tu, Shuming Shi, Tong Zhang, Yvette Graham, and Qun Liu.
\newblock Translating pro-drop languages with reconstruction models.
\newblock In {\em Proceedings of the AAAI Conference on Artificial
  Intelligence}, volume~32, 2018.

\bibitem{wang2018learning}
Longyue Wang, Zhaopeng Tu, Andy Way, and Qun Liu.
\newblock Learning to jointly translate and predict dropped pronouns with a
  shared reconstruction mechanism.
\newblock In {\em Proceedings of the 2018 Conference on Empirical Methods in
  Natural Language Processing}, pages 2997--3002, 2018.

\bibitem{koehn2007moses}
Philipp Koehn, Hieu Hoang, Alexandra Birch, Chris Callison-Burch, Marcello
  Federico, Nicola Bertoldi, Brooke Cowan, Wade Shen, Christine Moran, Richard
  Zens, et~al.
\newblock Moses: Open source toolkit for statistical machine translation.
\newblock In {\em Proceedings of the 45th Annual Meeting of the Association for
  Computational Linguistics Companion Volume Proceedings of the Demo and Poster
  Sessions}, pages 177--180. Association for Computational Linguistics, June
  2007.

\bibitem{zhang+:2020:texsmart}
Haisong Zhang, Lemao Liu, Haiyun Jiang, Yangming Li, Enbo Zhao, Kun Xu, Linfeng
  Song, Suncong Zheng, Botong Zhou, Jianchen Zhu, et~al.
\newblock Texsmart: A text understanding system for fine-grained ner and
  enhanced semantic analysis.
\newblock {\em arXiv preprint arXiv:2012.15639}, 2020.

\bibitem{sennrich+:2016}
Rico Sennrich, Barry Haddow, and Alexandra Birch.
\newblock Neural machine translation of rare words with subword units.
\newblock In {\em Proceedings of the 54th Annual Meeting of the Association for
  Computational Linguistics (Volume 1: Long Papers)}, pages 1715--1725, Berlin,
  Germany, August 2016. Association for Computational Linguistics.

\bibitem{papineni2002bleu}
Kishore Papineni, Salim Roukos, Todd Ward, and Wei-Jing Zhu.
\newblock Bleu: a method for automatic evaluation of machine translation.
\newblock In {\em ACL}, 2002.

\bibitem{domhan2018much}
Tobias Domhan.
\newblock How much attention do you need? a granular analysis of neural machine
  translation architectures.
\newblock In {\em ACL}, 2018.

\bibitem{Vaswani:2017:NIPS}
Ashish Vaswani, Noam Shazeer, Niki Parmar, Jakob Uszkoreit, Llion Jones,
  Aidan~N Gomez, {\L}ukasz Kaiser, and Illia Polosukhin.
\newblock Attention is all you need.
\newblock In {\em NeurIPS}, 2017.

\bibitem{wu2019pay}
Felix Wu, Angela Fan, Alexei Baevski, Yann~N Dauphin, and Michael Auli.
\newblock Pay less attention with lightweight and dynamic convolutions.
\newblock In {\em ICLR}, 2019.

\bibitem{ott:2019:naacl}
Myle Ott, Sergey Edunov, Alexei Baevski, Angela Fan, Sam Gross, Nathan Ng,
  David Grangier, and Michael Auli.
\newblock Fairseq: A fast, extensible toolkit for sequence modeling.
\newblock In {\em NAACL (Demonstrations)}, 2019.

\bibitem{ott2018scaling}
Myle Ott, Sergey Edunov, David Grangier, and Michael Auli.
\newblock Scaling neural machine translation.
\newblock In {\em WMT}, 2018.

\bibitem{Koehn2004:emnlp}
Philipp Koehn.
\newblock {Statistical Significance Tests for Machine Translation Evaluation}.
\newblock In {\em EMNLP}, 2004.

\bibitem{neubig2019:naacl}
Graham Neubig, Zi-Yi Dou, Junjie Hu, Paul Michel, Danish Pruthi, and Xinyi
  Wang.
\newblock compare-mt: A tool for holistic comparison of language generation
  systems.
\newblock In {\em NAACL (Demonstrations)}, 2019.

\bibitem{huang2018input}
Guoping Huang, Jiajun Zhang, Yu~Zhou, and Chengqing Zong.
\newblock Input method for human translators: A novel approach to integrate
  machine translation effectively and imperceptibly.
\newblock {\em ACM Transactions on Asian and Low-Resource Language Information
  Processing (TALLIP)}, 18(1):1--22, 2018.

\bibitem{steinberger2006jrc}
Ralf Steinberger, Bruno Pouliquen, Anna Widiger, Camelia Ignat, Toma{\v{z}}
  Erjavec, Dan Tufi{\c{s}}, and D{\'a}niel Varga.
\newblock The {JRC}-{A}cquis: A multilingual aligned parallel corpus with 20+
  languages.
\newblock In {\em Proceedings of the Fifth International Conference on Language
  Resources and Evaluation ({LREC}{'}06)}, Genoa, Italy, May 2006. European
  Language Resources Association (ELRA).

\bibitem{dinu2019training}
Georgiana Dinu, Prashant Mathur, Marcello Federico, and Yaser Al-Onaizan.
\newblock Training neural machine translation to apply terminology constraints.
\newblock In {\em Proceedings of the 57th Annual Meeting of the Association for
  Computational Linguistics}, pages 3063--3068, Florence, Italy, July 2019.
  Association for Computational Linguistics.

\bibitem{song-etal-2019-code}
Kai Song, Yue Zhang, Heng Yu, Weihua Luo, Kun Wang, and Min Zhang.
\newblock Code-switching for enhancing {NMT} with pre-specified translation.
\newblock In {\em Proceedings of the 2019 Conference of the North {A}merican
  Chapter of the Association for Computational Linguistics: Human Language
  Technologies}, pages 449--459, Minneapolis, Minnesota, June 2019. Association
  for Computational Linguistics.

\bibitem{klein2017opennmt}
Guillaume Klein, Yoon Kim, Yuntian Deng, Jean Senellart, and Alexander Rush.
\newblock {O}pen{NMT}: Open-source toolkit for neural machine translation.
\newblock In {\em Proceedings of the 55th Annual Meeting of the Association for
  Computational Linguistics}, pages 67--72, Vancouver, Canada, July 2017.
  Association for Computational Linguistics.

\bibitem{bialecki2012apache}
Andrzej Bia{\l}ecki, Robert Muir, Grant Ingersoll, and Lucid Imagination.
\newblock Apache lucene 4.
\newblock In {\em SIGIR 2012 workshop on open source information retrieval},
  page~17, 2012.

\bibitem{xu-etal-2020-boosting}
Jitao Xu, Josep Crego, and Jean Senellart.
\newblock Boosting neural machine translation with similar translations.
\newblock In {\em Proceedings of the 58th Annual Meeting of the Association for
  Computational Linguistics}, pages 1580--1590, 2020.

\bibitem{khandelwal2020nearest}
Urvashi Khandelwal, Angela Fan, Dan Jurafsky, Luke Zettlemoyer, and Mike Lewis.
\newblock Nearest neighbor machine translation.
\newblock {\em arXiv preprint arXiv:2010.00710}, 2020.

\bibitem{he2021fast}
Qiuxiang He, Guoping Huang, Qu~Cui, Li~Li, and Lemao Liu.
\newblock Fast and accurate neural machine translation with translation memory.
\newblock In {\em Proceedings of the 59th Annual Meeting of the Association for
  Computational Linguistics}, 2021.

\bibitem{cai2021neural}
Deng Cai, Yan Wang, Huayang Li, Wai Lam, and Lemao Liu.
\newblock Neural machine translation with monolingual translation memory.
\newblock In {\em Proceedings of the 59th Annual Meeting of the Association for
  Computational Linguistics}, 2021.

\bibitem{vaswani2018tensor2tensor}
Ashish Vaswani, Samy Bengio, Eugene Brevdo, Francois Chollet, Aidan~N Gomez,
  Stephan Gouws, Llion Jones, {\L}ukasz Kaiser, Nal Kalchbrenner, Niki Parmar,
  et~al.
\newblock Tensor2tensor for neural machine translation.
\newblock {\em arXiv preprint arXiv:1803.07416}, 2018.

\end{thebibliography}
\bibliographystyle{arxiv.sty}

\end{CJK}
\end{document}